\documentclass{article}

\PassOptionsToPackage{numbers, sort&compress}{natbib}

\usepackage[final,main]{neurips_2026}

\usepackage[utf8]{inputenc} 
\usepackage[T1]{fontenc}    
\usepackage[pagebackref,breaklinks,colorlinks]{hyperref}       
\usepackage{url}            
\usepackage{booktabs}       
\usepackage{array}          
\usepackage{amsfonts}       
\usepackage{amsmath}        
\usepackage{nicefrac}       
\usepackage{microtype}      
\usepackage{xcolor}         
\usepackage{graphicx}       
\usepackage{mwe}            
\usepackage{algorithm}      
\usepackage{algpseudocode}  
\usepackage{amssymb}        

\title{ProxySearch: Decoupled Inference-Time Scaling for Diffusion Models via Asymmetric Noise-Rank Transfer}

\author{%
  Seungwook Kim$^{1}$ \quad
  Jongmin Lee$^{2}$ \\[0.25cm]
  $^{1}$GenGenAI, 
  $^{2}$
  Chung-Ang University 
  \\[0.15cm]
  \texttt{wookiekim@gengen.ai} \quad
  \texttt{jmlee@cau.ac.kr}
}

\begin{document}

\maketitle

\begin{figure}[ht]
\vspace{-10mm}
  \centering
  \includegraphics[width=\linewidth]{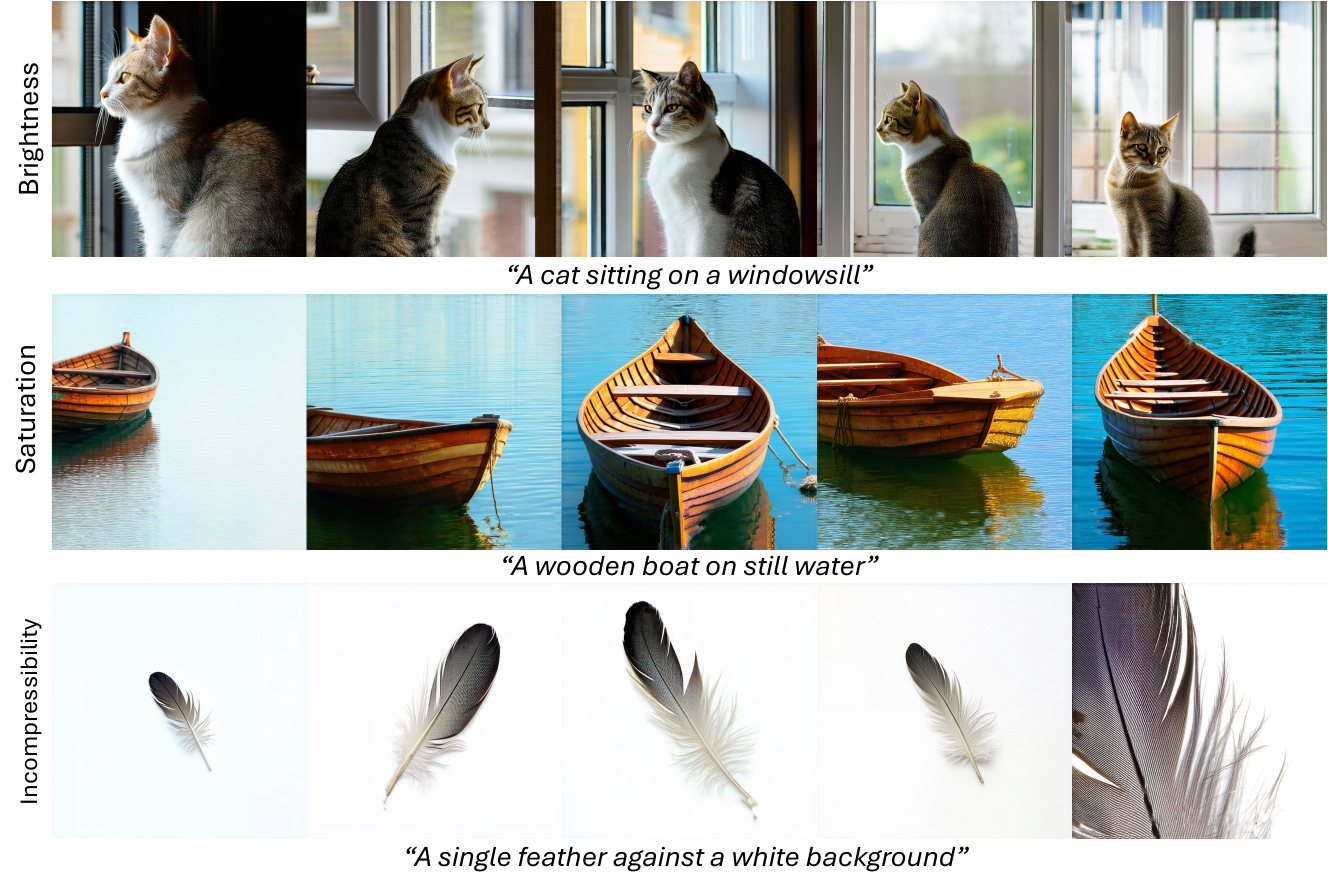}
  \vspace{-8mm}
  \caption{%
    \textbf{\textsc{ProxySearch} sets an image statistic by choosing the initial noise.}
    Each row is one SD3.5 prompt, with the target value of one statistic increasing from left to right.
  }
  \label{fig:teaser}
  \vspace{-2mm}
\end{figure}

\begin{abstract}
Noise search at inference time improves text-to-image diffusion but requires many multi-step teacher rollouts per query.
Distilled few-step models are now routinely released, yet their potential as \emph{proxies} inside noise search is uncharacterized.
We measure proxy--teacher rank agreement across three distillation regimes and three backbones (DiT, MMDiT, UNet) and find a sharp asymmetry.
On our main pool, closed-form image statistics transfer from the proxy to the teacher (Spearman $\rho$ up to $0.80$), while learned semantic scorers do not ($\rho\!\le\!0.12$).
Paired null controls trace this rank to the first denoising step for coarse statistics and to distillation for finer ones.
We use the asymmetry for noise search under a \emph{joint constraint}: a semantic reward specifies content, and a numeric target specifies image statistics.
In \textbf{\textsc{ProxySearch}}, the proxy filters candidates by statistical match in $1$ step, and the teacher returns the survivor with the highest semantic reward. No weight trades one objective against the other.
On FLUX-$1024$, \textsc{ProxySearch} tightens the statistical target by $21$--$23\%$ at the same number of teacher trajectories, with no detected loss on the optimized reward.
\end{abstract}

\section{Introduction}
\label{sec:intro}

Besides training better denoisers, text-to-image diffusion quality can also be improved by spending more compute at inference time, much as test-time compute is used in large language models~\citep{snell2024scaling}.
\emph{Noise search} draws many initial latents, renders each with the teacher denoiser, and keeps the highest-reward output~\citep{ma2025inference}. Refinements re-weight candidates along the denoising trajectory~\citep{singhal2024fksteering,wu2024practical}, optimize the noise directly~\citep{tang2024dno}, or learn a noise prior~\citep{zhou2024golden}.
In all of them, quality grows with the number of teacher rollouts per query. Every extra candidate costs a full multi-step rollout, so search cost grows linearly with the pool size.

Yet most major teachers now ship with a distilled few-step companion (FLUX-schnell, SD3.5-Large-Turbo, SDXL-Turbo~\citep{sauer2024ladd,sauer2023adversarial}) that needs $1$--$4$ instead of $28$--$50$ steps.
Search pipelines use this sibling only as a faster generator, and whether it can also judge candidates has not been tested. Inside search, a proxy helps only if it ranks candidates as the teacher would; producing similar images on average is not enough.
We measure, for each verifier, whether a distilled sibling ranks a prompt's candidate noises as its teacher does. The answer depends on the verifier (Fig.~\ref{fig:transfer}). Eight closed-form image statistics, such as brightness and contrast, transfer from proxy to teacher (Spearman $\rho$ up to $0.80$ on our main pool), while three learned semantic rewards (CLIP~\citep{radford2021clip}, ImageReward~\citep{xu2023imagereward}, aesthetic~\citep{schuhmann2022aesthetic}) stay at $\rho\!\le\!0.12$ on the same noises. On every pair, rank agreement on brightness is $6$--$16\times$ that on CLIP score.
Paired null controls show that the raw noise does not carry this rank. A single step of the \emph{undistilled} teacher already ranks coarse statistics such as brightness, and distillation extends rank agreement to finer ones such as edge density. The distilled proxy is the only $1$-step model we test that ranks all eight. So the cheap proxy can screen many candidates on image statistics, while semantics still have to be judged by the teacher.

\begin{figure}[t]
  \centering
  \includegraphics[width=0.95\linewidth]{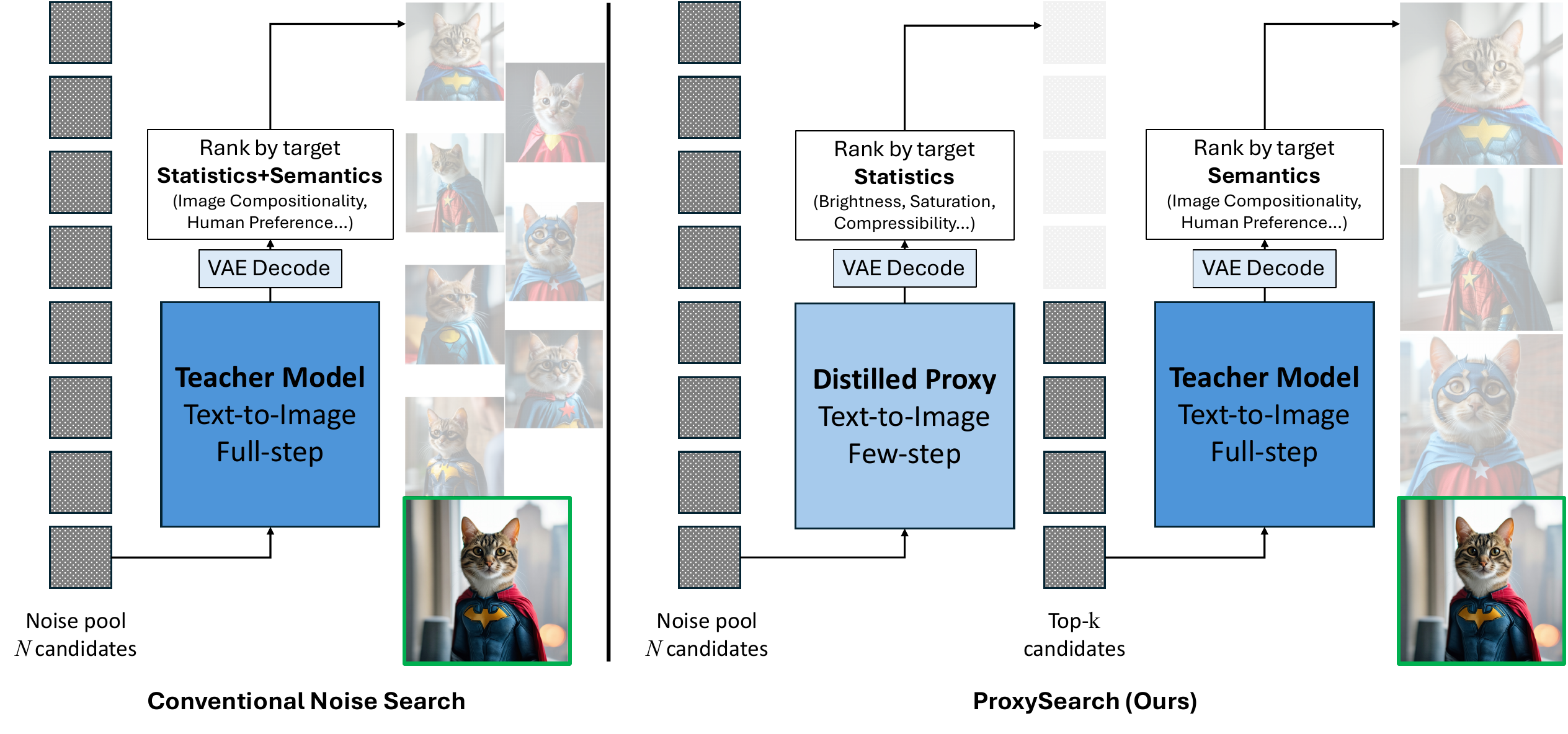}
  \vspace{-5.2mm}
  \caption{%
    \textbf{Conventional noise search vs.\ \textsc{ProxySearch}.}
    Conventional search renders all $N$ candidates with the teacher $\mathcal{M}$ and keeps the highest-reward one.
    \textsc{ProxySearch} scores the same $N$ noises on $g_\mathrm{stat}$ with the distilled proxy $\tilde{\mathcal{M}}$ at about $1$ NFE each, renders only the top $K$ with the teacher, and returns the survivor with the highest semantic reward.
  }
  \vspace{-4mm}
  \label{fig:pipeline}
\end{figure}

A user may want a batch of images rendered to one brightness and contrast, or a synthetic image set balanced along a statistic~\citep{nam2026training}. Choosing the noise changes such a look (Fig.~\ref{fig:teaser}), but naming the look in the prompt does not set it (Fig.~\ref{fig:qualitative-prompt-spec}). We therefore cast search as a \emph{joint constraint}. The prompt and a semantic reward, such as compositional accuracy or learned preference, specify content. A separate numeric target $\tau$, such as a brightness and contrast level, specifies image statistics. Search should return an output that scores well on the semantic reward \emph{and} hits $\tau$, without scalarizing the two.
\textsc{ProxySearch} (Fig.~\ref{fig:pipeline}) routes each part of the constraint to the model that ranks it. The proxy filters $N$ candidates by statistical match at about one NFE each, and the teacher renders the top $K$ and returns the one with the highest semantic reward. The method has two integer parameters, $N$ and $K$, and no weight trading one objective against the other.

On full GenEval at FLUX-$1024$ (Tab.~\ref{tab:hero}), \textsc{ProxySearch} runs the same $8$ teacher trajectories as teacher-only argmax ($+16\%$ denoising NFE). It tightens paired statistical $\ell_1$ by $21$--$23\%$ under three search rewards (paired $t\!=\!-19.3$ under GenEval) and stays on par with teacher argmax on the reward it optimizes.
More teacher candidates do not produce this gain: under semantic-only selection, $16$- and $64$-candidate teacher pools leave $\ell_1$ unchanged (Sec.~\ref{sec:exp:headline}). The gain therefore comes from screening on the statistic, not from a larger candidate pool. On a $100$-cell pilot, a composite reward that scores all $64$ teacher renders reaches a lower $\ell_1$ than \textsc{ProxySearch}, but at $6.9\times$ the total NFE (Sec.~\ref{sec:exp:alternatives}).
The tightening also replicates in $50$-prompt pilots on SD3.5~\citep{esser2024scaling} and SDXL~\citep{podell2023sdxl}, at a GenEval cost of $2$--$4$~pp relative to teacher-only argmax.

Our contributions are as follows:
\begin{itemize}
  \item A characterization of proxy-to-teacher rank agreement on released distilled pairs, which separates statistics from semantics. On our main pool, closed-form image statistics transfer with Spearman $\rho$ up to $0.80$, while learned semantic rewards remain at $\rho\!\le\!0.12$. Paired null controls show that a single undistilled teacher step already ranks coarse statistics and that distillation extends rank agreement to finer ones (Sec.~\ref{sec:method:transfer}).
  \item \textsc{ProxySearch}, a reformulation of inference-time noise search as a joint constraint on a statistical target and a user-chosen semantic reward, parameterized by two integers ($N$, $K$) with no trade-off weight to tune (Sec.~\ref{sec:method}).
  \item Empirical evidence that \textsc{ProxySearch} tightens the statistical target by $21$--$23\%$ in paired $\ell_1$ on the full GenEval suite at FLUX-$1024$, at the same number of teacher trajectories as teacher-only argmax, while staying on par on the optimized reward. Pilots on SD3.5 and SDXL replicate the tightening (Sec.~\ref{sec:exp}).
\end{itemize}

\section{Related Work}
\label{sec:related}

\smallbreak
\noindent
\textbf{Inference-time scaling for diffusion.}
A growing line of work treats the initial noise as a search variable. \citet{ma2025inference} show that random search, zero-order optimization, and path-level search all keep improving as verifier compute grows. Particle-based variants~\citep{wu2024practical,singhal2024fksteering} resample partial trajectories, gradient-based variants~\citep{tang2024dno,zhou2024golden,eyring2024reno} optimize the latent or learn a noise prior, and initialization and trajectory-stylizing methods~\citep{guo2024initno,bai2025zigzag,kim2025das,kim2025freeaction} steer the search distribution itself. Reward-guided and restart-style sampling~\citep{bansal2023universalguidance,yu2023freedom,xu2023restart,karras2024autoguidance}, building on classifier-free guidance~\citep{ho2022classifierfree}, similarly trades teacher NFE for higher-reward outputs. Sampling- and optimization-time techniques also improve generation beyond single images, in video~\citep{kim2025freeaction} and text-to-3D~\citep{kim2024correspondentdream,lee2025harnessing}, and multi-view diffusion models~\citep{kim2024mvip,kim2025rapidmv} generate consistent views for 3D generation. The search methods above score every candidate on teacher outputs and treat the reward as a single scalar. We instead split the reward into statistics, which transfer from a distilled proxy at little cost, and semantics, which do not.

\smallbreak
\noindent
\textbf{Distillation of diffusion models.}
A parallel literature compresses the $28$--$50$-step teacher schedule~\citep{ho2020denoising,song2021denoising,karras2022elucidating,lipman2023flow} into $1$--$4$ steps, through progressive distillation~\citep{salimans2022progressive}, consistency models~\citep{song2023consistency,luo2023lcmlora}, rectified-flow and one-step variants~\citep{liu2022rectified,liu2024instaflow}, adversarial distillation (LADD~\citep{sauer2024ladd}, which produces FLUX-schnell, and the related ADD~\citep{sauer2023adversarial} behind SD3.5-Large-Turbo and SDXL-Turbo), distribution matching~\citep[DMD/DMD2;][]{yin2024dmd,yin2024dmd2}, and Hyper-SD~\citep{ren2024hypersd}. This literature evaluates the distilled model as a standalone fast generator, reporting FID~\citep{heusel2017fid} or human preference against the teacher. To our knowledge, nobody has checked whether the proxy keeps the teacher's \emph{rank} of a candidate-level reward across noises drawn for the same prompt; we measure this in Sec.~\ref{sec:method:transfer}.

\smallbreak
\noindent
\textbf{Cheap proxies for expensive search.}
Pairing a cheap, biased estimator with an expensive accurate one recurs across machine learning: multi-fidelity Bayesian optimization~\citep{kandasamy2017multifidelity}, Hyperband-style allocation~\citep{li2018hyperband,falkner2018bohb}, AlphaGo's value-network/rollout-policy split~\citep{silver2016alphago}, one-shot NAS~\citep{pham2018enas}, speculative decoding~\citep{leviathan2023speculative,stern2018blockwise,chen2023speculative,cai2024medusa}, learned process verifiers~\citep{cobbe2021training,lightman2024verify,brown2024monkeys}, and cascaded diffusion~\citep{ho2022cascaded}. In each, the proxy is \emph{designed} to work with the full evaluator.
Diffusion distillation provides no such guarantee. The distilled sibling is trained to produce teacher-quality samples in few steps, not to preserve the teacher's ranking of inputs, and for the same noise the two models produce \emph{different} images. Nor does the cascade pattern predict that rank agreement would split along statistics and semantics. We instead measure where it appears: on the statistical axes (Sec.~\ref{sec:method:transfer}).

\smallbreak
\noindent
\textbf{Reward fine-tuning of diffusion.}
A parallel paradigm bakes reward signals into the model rather than into the sampler: by backpropagating a differentiable reward (DRaFT~\citep{clark2023draft}, AlignProp~\citep{prabhudesai2023alignprop}, TextCraftor~\citep{li2024textcraftor}), by policy gradients on the denoising MDP (DPOK~\citep{fan2023dpok}, DDPO~\citep{black2023ddpo}), possibly with intrinsic self-confidence rewards instead of external reward models~\citep{kim2026solace}, by preference optimization (Diffusion-DPO~\citep{wallace2024diffusiondpo} and follow-ups~\citep{yang2024d3po,liang2025spo}), or by rejection sampling and adjoint matching~\citep{lee2023aligning,dong2024raft,domingoenrich2024adjoint}. These methods produce a single tuned model per reward. \textsc{ProxySearch} is orthogonal: it is training-free and leaves both checkpoints untouched. A reward-post-trained teacher still agrees with a fixed distilled proxy on statistical rank (Sec.~\ref{sec:exp:rl-postrain}).

\section{Preliminaries}
\label{sec:prelim}

\paragraph{Diffusion sampling and noise search.}
A diffusion model $\mathcal{M}$~\citep{ho2020denoising,song2021denoising,song2021scorebased,dhariwal2021beats,karras2022elucidating} maps an initial latent $z_T \sim \mathcal{N}(0,I)$ and a text prompt $c$ to a latent $z_0 = \mathcal{M}_T^{\to 0}(z_T, c)$, decoded by a VAE to image $x$~\citep{rombach2022high}. Inference-time noise search draws $N$ noises $\{z_T^{(i)}\}_{i=1}^{N}$, renders $x^{(i)} = \mathrm{VAE}(\mathcal{M}(z_T^{(i)},c))$, scores each with a verifier $v(\cdot)$, and returns $x^{(i^{\ast})}$ with $i^{\ast}\!=\!\arg\max_i v(x^{(i)})$. Cost is dominated by the $N$ teacher evaluations, each requiring $28$--$50$ denoising steps. We count cost in neural function evaluations (NFE), one per denoising step. NFE excludes classifier-free-guidance passes, VAE decoding, text encoding, and verifier scoring, and Sec.~\ref{sec:exp:impl} reports measured render time.

\paragraph{Distilled proxy.}
A distilled variant $\tilde{\mathcal{M}}$ (FLUX-schnell, SD3.5-Large-Turbo, and SDXL-Turbo~\citep{sauer2024ladd,sauer2023adversarial}) shares the architecture, tokenizer, VAE, and latent space of $\mathcal{M}$ but produces an output in $k_\mathrm{proxy}$ steps. We use $k_\mathrm{proxy}\!=\!1$ unless noted, and Tab.~\ref{tab:appendix:exp-protocol} lists the setting of every run. We write $\tilde{x}^{(i)} = \mathrm{VAE}(\tilde{\mathcal{M}}(z_T^{(i)},c))$ for the proxy image on the \emph{same} noise and prompt.

\paragraph{Two verifier families.}
We partition the verifiers used in this paper into two families:
\begin{itemize}
  \item \textbf{Statistics:} model-free global image statistics: brightness, contrast, saturation, colorfulness (Hasler--S\"usstrunk~\citep{hasler2003colorfulness}), sharpness (Laplacian variance), compressibility (JPEG size), edge density (Sobel~\citep{sobel1970camera}), and color temperature. Each costs $\ll$ one teacher NFE.
  \item \textbf{Semantics:} prompt-dependent, model-based semantic scores. The rank-transfer bank uses CLIP similarity~\citep{radford2021clip}, ImageReward~\citep{xu2023imagereward}, and a LAION aesthetic predictor~\citep{schuhmann2022aesthetic}; GenEval pass~\citep{ghosh2023geneval}, PickScore~\citep{kirstain2023pickscore}, and HPSv2~\citep{wu2023hpsv2} serve as search objectives but are not part of the rank-transfer measurement.
\end{itemize}
Exact definitions, units, and the GenEval selection and tie-breaking rules are given in App.~\ref{sec:appendix:details}.

\paragraph{Statistical target.}
A user supplies, separately from the prompt, a target vector $\tau\!\in\!\mathbb{R}^{d}$ over a chosen subset of $d$ statistical axes, either directly or via a preset. We measure adherence by the negative $\ell_1$ distance $g_\mathrm{stat}(\ell;\tau)\!=\!-\|\ell-\tau\|_1$, where $\ell\!=\!\mathrm{stat}(x)$ is the statistics vector of image $x$ restricted to the $d$ axes named by $\tau$. For example, a ``bright, high-contrast'' look may correspond to $\tau\!=\!(\mathrm{brightness}\!=\!0.72,\,\mathrm{contrast}\!=\!0.26)$. The target $\tau$ is a per-query specification, not a model parameter, and our main experiments use two fixed targets (Sec.~\ref{sec:exp:headline}).

\section{Distilled proxies rank image statistics, not learned semantic scores}
\label{sec:transfer}
\label{sec:method:transfer}
\label{sec:exp:transfer}

\begin{figure}[t]
  \centering
  \includegraphics[width=1.0\linewidth]{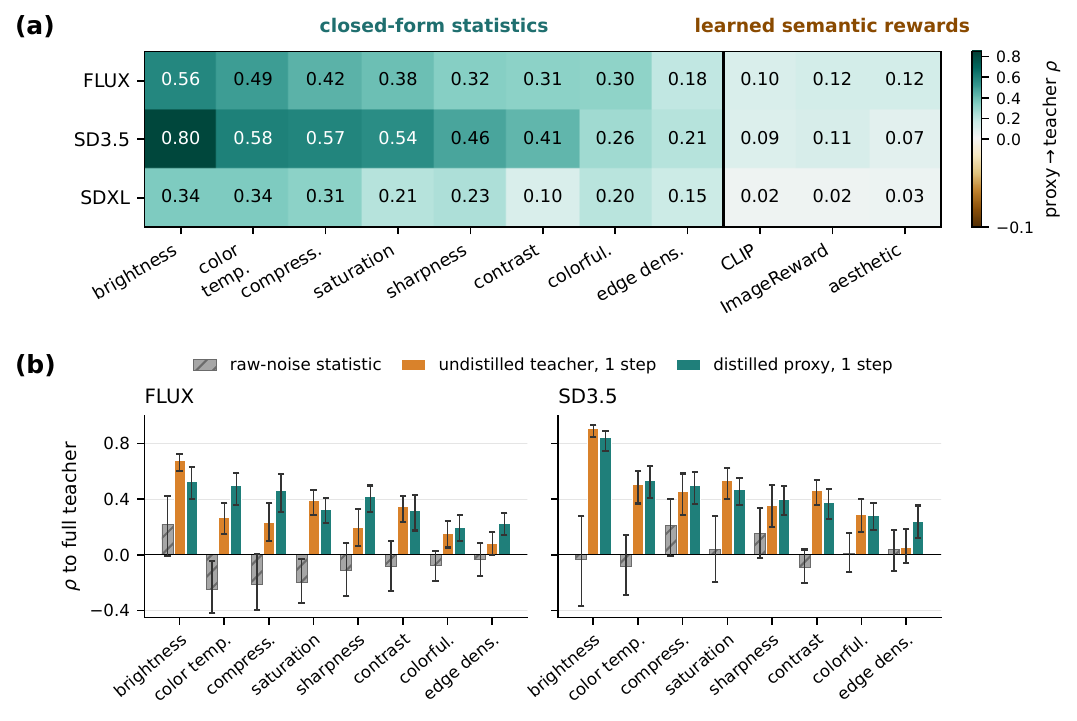}
  \vspace{-6mm}
  \caption{%
    \textbf{Proxy--teacher rank correlation splits along statistics vs.\ semantics.}
    \textbf{(a)} Spearman $\rho$ between the proxy's and the teacher's rankings of the same noises, for $11$ verifiers on $3$ model pairs. Statistics transfer with $\rho$ up to $0.80$, while semantics stay at $\rho\!\le\!0.12$.
    \textbf{(b)} Paired null controls on a shared noise pool. Raw-noise statistics do not transfer, one undistilled teacher step ranks coarse statistics, and the distilled proxy also ranks finer ones (most clearly on FLUX). It is the only condition significant on all eight.%
  }
  \label{fig:transfer}
  \vspace{-5.5mm}
\end{figure}

\smallbreak
\noindent
\textbf{Protocol and the split.}
For each prompt and each of $32$ noises, we render the proxy image $\tilde{x}^{(i)}$ ($k_\mathrm{proxy}\!=\!1$) and the teacher image $x^{(i)}$ at $512^{2}$ and score both with $11$ verifiers ($8$ statistics, $3$ semantics). For each verifier, we then compute the Spearman correlation between the proxy and teacher rankings of the $32$ noises of a prompt. We average this per-prompt $\rho$ over $100$ synthetic template prompts (App.~\ref{sec:appendix:details}) on the three pairs of Tab.~\ref{tab:protocol}: FLUX (DiT, LADD~\citep{sauer2024ladd}), SD3.5 (MMDiT~\citep{esser2024scaling}, ADD~\citep{sauer2023adversarial}), and SDXL (UNet~\citep{podell2023sdxl}, ADD). Fig.~\ref{fig:transfer}~(a) shows the result.
On our main pool, statistical verifiers transfer at $\rho\!\in\![0.10, 0.80]$ and semantic verifiers stay at $\rho\!\le\!0.12$. 
Strength varies across axes and backbones, but on every pair rank agreement on brightness is $6$--$16\times$ that on CLIP score. The proxy ranks image statistics but barely ranks semantics. Selecting candidates by proxy-side semantic scores stays far below teacher-side selection (Tab.~\ref{tab:appendix:proxy-hf-ceiling}). The labels ``statistics'' and ``semantics'' describe the eight statistics and three semantic scorers we measure. App.~\ref{sec:appendix:limits} discusses other verifiers.

\smallbreak
\noindent
\textbf{Where the statistical rank comes from.}
We run three conditions on one shared pool of $12$ GenEval prompts $\times$ $32$ noises at $1024^{2}$, so that noise $j$ is the identical $z_T$ for every model. Each condition is scored against the full teacher (Fig.~\ref{fig:transfer}~(b), Tab.~\ref{tab:appendix:paired-null}). Statistics of the raw initial latent do not transfer at all. A single step of the undistilled teacher already preserves coarse-statistic rank (brightness $\rho\!=\!0.68$ on FLUX, $0.91$ on SD3.5). Distillation adds rank agreement on finer statistics where the $1$-step teacher is weaker. On FLUX it raises color temperature $0.27\!\to\!0.50$, compressibility $0.23\!\to\!0.46$, sharpness $0.20\!\to\!0.42$, and edge density $0.08\!\to\!0.23$. On SD3.5 the gain is confined to edge density ($0.06\!\to\!0.24$). The distilled proxy is the only $1$-step condition with significant rank agreement on all eight statistics. Two further controls indicate that the statistic a teacher produces from a given noise is largely model-specific (App.~\ref{sec:appendix:shared-noise}). Two different teachers rendering the same $z_T$ agree only weakly (mean $|\rho|\!=\!0.07$ over the eight statistics), with a weak shared brightness component. A predictor trained on raw $z_T$ recovers part of a teacher's brightness rank within a model but transfers little across models.

The first prompt-conditioned denoising step thus seeds coarse statistical rank, and distillation extends it to finer statistics. This fits the coarse-to-fine view of denoising~\citep{rissanen2023idhp,dieleman2024spectral}, although our controls do not pin the effect down to one mechanism.

\begin{figure}[t]
  \centering
  \includegraphics[width=0.92\linewidth]{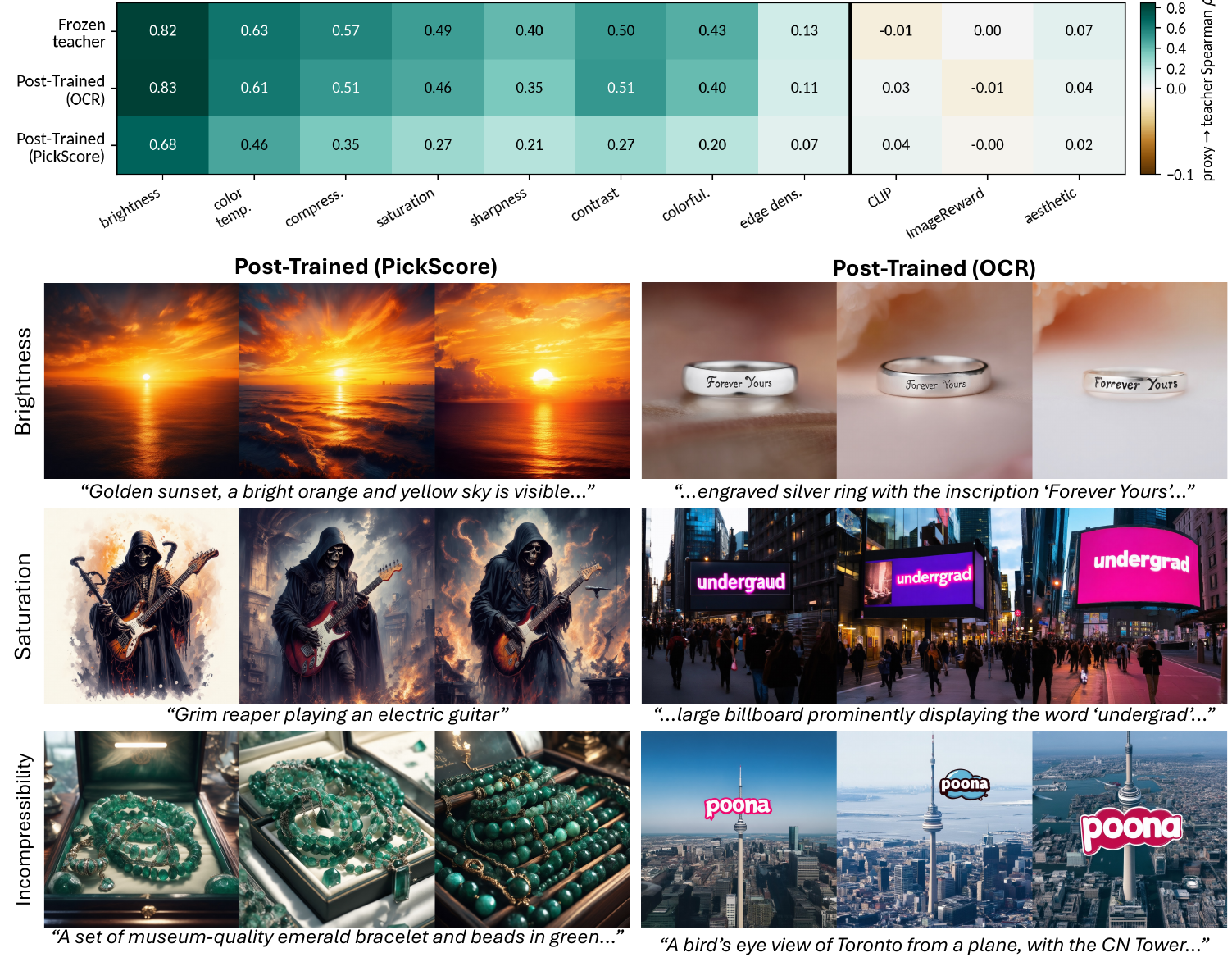}
  \vspace{-5.5mm}
  \caption{%
    \textbf{The statistics/semantics split persists under the two tested RL post-trainings of the teacher.}
    \emph{Top:} proxy$\to$teacher Spearman $\rho$ across $11$ verifiers (rows). 
    \emph{Bottom:} qualitative samples per statistical axis (brightness, saturation, incompressibility).
  }
  \label{fig:rl-postrain}
  \vspace{-3mm}
\end{figure}

\smallbreak
\noindent
\textbf{Robustness of the split.}\label{sec:exp:rl-postrain}
On FLUX and SD3.5, a $4$-step proxy does not recover semantic rank ($\rho\!<\!0.16$) and leaves brightness and color-temperature rank unchanged, at about $3\times$ the proxy cost. On SDXL it lowers most statistical axes (App.~\ref{sec:appendix:proxy-step-count}). A DMD2 score-distilled SDXL proxy~\citep{yin2024dmd2} shows the same split as the LADD/ADD proxies (App.~\ref{sec:appendix:per-arch}). The split therefore does not seem tied to one particular distillation loss.
Released teachers are also increasingly post-trained with reward RL~\citep{liu2025flowgrpo,wallace2024diffusiondpo,kim2026solace}, and a fixed proxy serves such a teacher only if the split survives the tuning. We pair one fixed proxy, a third-party distilled SD3.5-Medium model run at $1$ step, with the SD3.5-Medium teacher in three versions: frozen, with a FlowGRPO text-rendering LoRA, and with a FlowGRPO PickScore LoRA. Each arm uses $100$ FlowGRPO prompts $\times$ $32$ noises at $1024^{2}$, and arms are paired by prompt only (App.~\ref{sec:appendix:rl-postrain}). The text-rendering LoRA stays within about $0.06$ of the frozen teacher on every statistical axis (Fig.~\ref{fig:rl-postrain}). The PickScore LoRA lowers statistical $\rho$ by roughly $0.14$--$0.23$ on every axis except edge density ($0.133\!\to\!0.068$). Brightness rank agreement ($0.68$) remains $15\times$ that of the best semantic verifier ($0.045$). Of the $50$ PickScore prompts, $48$ lie in that LoRA's training split, and other RL objectives or LoRA settings may weaken the split further.

\section{ProxySearch}
\label{sec:method}

\subsection{Problem setup: joint-constraint search}
\label{sec:method:problem}

\begin{figure}[h]
  \centering
  \includegraphics[width=0.95\linewidth]{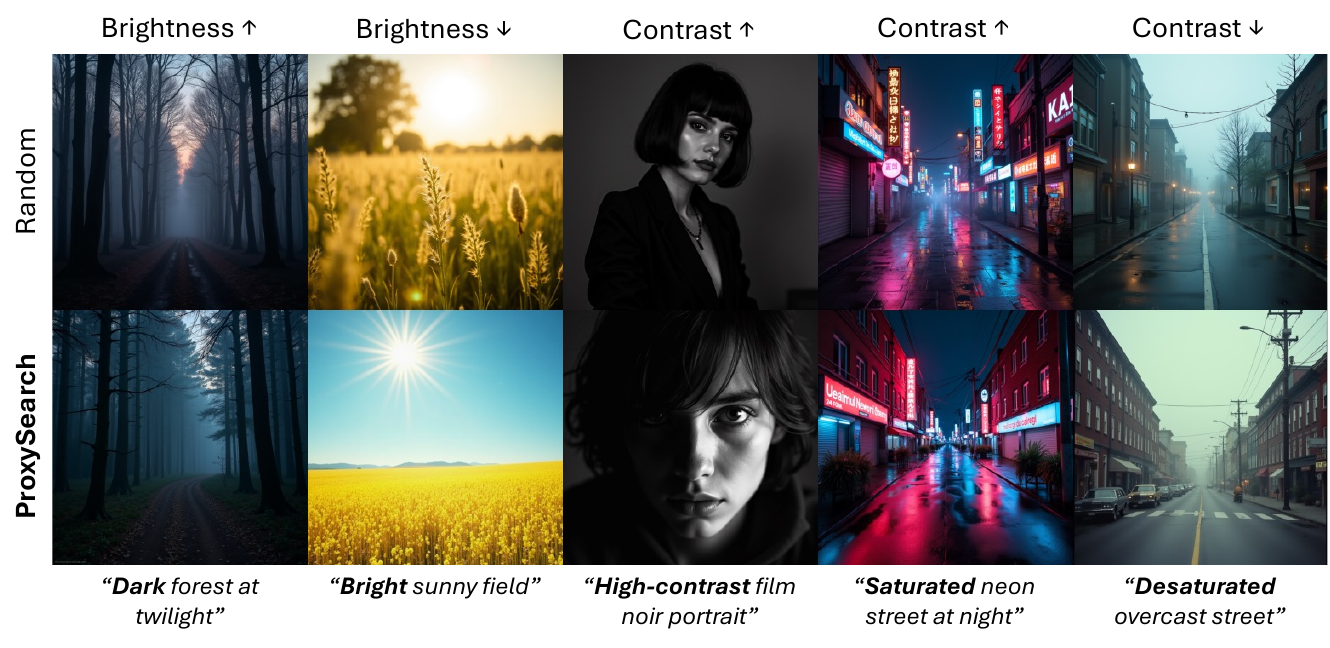}
  \vspace{-5mm}
  \caption{%
    \textbf{Prompt wording alone leaves wide variance on the named statistical axis.}
    Five prompts (columns) that already name a statistical axis directly (``dark'', ``bright'', ``high-contrast'', ``saturated'', ``desaturated'') still leave wide variance on the named axis across the $32$-noise output pool.
  }
  \label{fig:qualitative-prompt-spec}
  \vspace{-5mm}
\end{figure}

Prompt wording alone does not pin down numeric image statistics: even prompts that name a statistical axis directly (``dark'', ``bright'', ``saturated'') still vary widely on the named axis across noises (Fig.~\ref{fig:qualitative-prompt-spec}). Let $\mathcal{M}$ be a teacher diffusion model and $\tilde{\mathcal{M}}$ its distilled few-step proxy (Sec.~\ref{sec:prelim}). Given a prompt $c$, a user-specified semantic reward $v_\mathrm{sem}$ such as GenEval pass, PickScore, or HPSv2, and a statistical target $\tau$ (Sec.~\ref{sec:prelim}), the task is to select $x^{\ast}$ from a noise pool $\{z_T^{(i)}\}_{i=1}^{N}$ that satisfies both criteria. An idealized form of the task is the constrained argmax $\max_i v_\mathrm{sem}(x^{(i)})$ subject to $g_\mathrm{stat}(\mathrm{stat}(x^{(i)});\tau)\!\ge\!-\epsilon$ for a user tolerance $\epsilon$, which \textsc{ProxySearch} approximates with a top-$K$ heuristic (Sec.~\ref{sec:method:algorithm}). All reported experiments use the target-matching $g_\mathrm{stat}$ of Sec.~\ref{sec:prelim}. The algorithm applies to any $g_\mathrm{stat}$ for which larger values are preferred (App.~\ref{sec:appendix:details}).

The two criteria are conventionally handled in one of two ways: (a) collapsed into a single scalarized reward $v_\mathrm{sem}(x) + \lambda\,g_\mathrm{stat}(\mathrm{stat}(x);\tau)$ and maximized by teacher-only search, or (b) handled on the teacher alone, either by post-processing the teacher-argmax image toward $\tau$ or by filtering teacher renders on $g_\mathrm{stat}$ before the semantic argmax. Neither route uses the proxy, although Sec.~\ref{sec:method:transfer} shows that it ranks statistics. Route (a) scores both terms on teacher renders under a tuned weight $\lambda$. Route (b) spends teacher rollouts on candidates whose statistics could have been screened cheaply, and post-processing reaches $\tau$ only by editing pixels. We instead treat the two as a \emph{joint constraint}: statistics are screened on cheap proxy renders, and semantics are scored only on teacher renders, the images actually returned. Sec.~\ref{sec:exp} compares these baselines with \textsc{ProxySearch}.

\subsection{Algorithm and cost}
\label{sec:method:algorithm}

\textsc{ProxySearch} routes the statistical part of the constraint to the proxy and the semantic part to the teacher (Fig.~\ref{fig:pipeline}). Given a noise budget $N$ and a survivor budget $K$:
\begin{enumerate}
  \item \textbf{Proxy statistical filter.} For each $z_T^{(i)},\,i\!=\!1,\dots,N$, render
        $\tilde{x}^{(i)}=\mathrm{VAE}(\tilde{\mathcal{M}}(z_T^{(i)},c))$ at
        $k_\mathrm{proxy}$ proxy steps ($k_\mathrm{proxy}\!=\!1$ unless noted).
        Score the statistical objective
        $\ell^{(i)}=g_\mathrm{stat}(\mathrm{stat}(\tilde{x}^{(i)});\tau)$.
        Retain the indices $\mathcal{S}\!\subseteq\![N]$ with the $K$
        largest $\ell^{(i)}$.
  \item \textbf{Teacher semantic argmax.} For each $i\!\in\!\mathcal{S}$, render
        $x^{(i)}=\mathrm{VAE}(\mathcal{M}(z_T^{(i)},c))$ at the teacher's
        full step count. Return $x^{(i^{\ast})}$ with
        $i^{\ast}\!=\!\arg\max_{i\in\mathcal{S}} v_\mathrm{sem}(x^{(i)})$.
\end{enumerate}
Teacher renders are not re-scored on $g_\mathrm{stat}$, so the returned image carries no tolerance guarantee. The survivors are the $K$ best proxy candidates even when $\tau$ is not attainable in the pool. \textsc{ProxySearch} has no weight $\lambda$, but it still trades the two objectives. A smaller $K/N$ screens harder on statistics and leaves fewer candidates for the semantic argmax. The attainability of $\tau$ in the pool and the scaling of the axes inside $g_\mathrm{stat}$ also shift the balance. The final argmax is optimal only over the survivor set $\mathcal{S}$ and the chosen $v_\mathrm{sem}$.

The pipeline costs $N\cdot c_\mathrm{proxy}+K\cdot c_\mathrm{teacher}$ NFE. With a $1$-step proxy, $c_\mathrm{proxy}/c_\mathrm{teacher}$ is $1/50$ on FLUX-$1024$ and SDXL ($50$ teacher steps) and $1/28$ on SD3.5 ($28$ teacher steps). At the FLUX-$1024$ operating point ($N\!=\!64$, $K\!=\!8$), \textsc{ProxySearch} runs the same $8$ teacher trajectories as an $8$-candidate teacher-only argmax and adds $64$ proxy NFE to the teacher's $400$. This is $16\%$ more denoising NFE. We call this matched \emph{teacher trajectories}, not matched total compute, and Sec.~\ref{sec:exp:impl} reports the measured render time.

\section{Experiments}
\label{sec:exp}

\subsection{Setup}
\label{sec:exp:protocol}
\label{sec:exp:impl}

We evaluate on FLUX.1-dev / FLUX-schnell (DiT, LADD~\citep{sauer2024ladd}), SD3.5-Large / SD3.5-Large-Turbo~\citep{esser2024scaling} (MMDiT, ADD), and SDXL / SDXL-Turbo~\citep{podell2023sdxl,sauer2023adversarial} (UNet, ADD). The compositional benchmark for image pairs is GenEval~\citep{ghosh2023geneval}. The human-preference rewards are PickScore~\citep{kirstain2023pickscore} and HPSv2~\citep{wu2023hpsv2}, and CLIP, ImageReward, and the LAION aesthetic predictor serve as held-out semantic measurements. Search experiments use GenEval prompts. GE denotes GenEval scored with the official detection pipeline, reported as the per-image fraction of satisfied prompt requirements averaged over cells.\footnote{GE-driven selection uses an in-process re-implementation of the GenEval checker (Mask2Former via HF \texttt{transformers}); reported GE re-scores the selected images with the official pipeline. Selection keys and tie-breaks are in App.~\ref{sec:appendix:details}.} Details in App.~\ref{sec:appendix:details}.

\smallbreak
\noindent
\textbf{Compute and pairing.}
At matched teacher trajectories, teacher-only argmax renders $N_t$ teacher candidates and \textsc{ProxySearch} keeps $K\!=\!N_t$ survivors, so both run $K$ teacher trajectories per query. \textsc{ProxySearch} adds $N\!\cdot\!k_\mathrm{proxy}$ proxy NFE: $464$ vs.\ $400$ NFE at the FLUX-$1024$ operating point ($+16\%$). Measured render time grows more than NFE because each proxy render adds a VAE decode and a text encoding: about $+24\%$ on FLUX ($186$ vs.\ $230$~s). App.~\ref{sec:appendix:compute} gives guidance passes and approximate render times for SD3.5 ($+28\%$) and SDXL ($+49\%$). Within a run, all methods share one noise pool indexed by $(\mathrm{seed},\,\mathrm{prompt\_id},\,\mathrm{candidate\_idx})$. Teacher argmax uses the first $N_t$ candidates of the pool that \textsc{ProxySearch} screens, so reported differences are paired. Pools are not shared across runs, so comparisons are paired only within a run. All main results use one fixed $K$, and Sec.~\ref{sec:exp:ksweep} sweeps it.

\subsection{\textsc{ProxySearch} tightens statistics at matched teacher trajectories}
\label{sec:exp:headline}

We run the full GenEval suite at FLUX-$1024$ with two statistical targets, $\tau_\mathrm{bright}\!=\!(\mathrm{brightness}, \mathrm{contrast})\!=\!(0.72, 0.26)$ and $\tau_\mathrm{dark}\!=\!(0.22, 0.20)$. This gives $n\!=\!1106$ paired prompt--target cells ($553$ prompts $\times$ $2$ targets). We compare \textsc{ProxySearch} ($N\!=\!64$, $K\!=\!8$, $k_\mathrm{proxy}\!=\!1$) against teacher argmax with $N_t\!=\!K$, so both methods render $8$ teacher trajectories per cell.
Tab.~\ref{tab:hero} reports three search objectives ($v_\mathrm{sem}\!\in\!\{$GE, PickScore, HPSv2$\}$). Every reward other than the search objective is a held-out post-selection measurement. Tab.~\ref{tab:hero} also lists \emph{Teacher-seq}, which spends $2\times$ the teacher NFE: it renders $N_t\!=\!16$ teacher candidates, keeps the $8$ with the smallest teacher-side statistical $\ell_1$, and argmaxes $v_\mathrm{sem}$ over them. Teacher-seq was measured on a paired $100$-cell subset ($50$ GenEval prompts $\times$ $2$ targets). Its rows are extrapolated to $n\!=\!1106$ as the \textsc{ProxySearch} value plus the paired Teacher-seq$-$\textsc{ProxySearch} difference on that subset.

\begin{table}[t]
  \centering
  \small
  \caption{%
    \textbf{\textsc{ProxySearch} tightens statistics while matching teacher argmax on the optimized semantic reward.} FLUX-$1024$, $N\!=\!64$, $K\!=\!N_t\!=\!8$, $k_\mathrm{proxy}\!=\!1$, $n\!=\!1106$ paired prompt--target cells (Teacher-seq rows extrapolated from a paired $100$-cell subset; see text), three search objectives: GE-driven (a); PickScore-driven (b); HPSv2-driven (c). Bold marks the best entry per column within each panel (ties all bold). Paired deltas to teacher argmax are in App.~\ref{sec:appendix:hero-ci}.
  }
  \label{tab:hero}
  \resizebox{\textwidth}{!}{%
  \begin{tabular}{lccccccc}
    \toprule
    Method & statistical $\ell_1\!\downarrow$ & GE \%$\uparrow$ & PickScore $\uparrow$ & HPSv2 $\uparrow$ & CLIP $\uparrow$ & ImageReward $\uparrow$ & aesthetic $\uparrow$ \\
    \midrule
    \multicolumn{8}{l}{\emph{(a) GenEval-driven search ($v_\mathrm{sem}\!=\!$\,GE)}} \\
    Teacher argmax ($N_t\!=\!8$) & 0.302 & 81.6 & 23.50 & 0.311 & 0.364 & 1.06 & \textbf{5.35} \\
    \textbf{\textsc{ProxySearch} ($K\!=\!8$)} & \textbf{0.233} & \textbf{82.5} & \textbf{23.52} & \textbf{0.312} & \textbf{0.365} & \textbf{1.12} & \textbf{5.35} \\
    Teacher-seq ($N_t\!=\!16$, $2\!\times$ NFE) & 0.247 & \textbf{82.5} & 23.49 & 0.307 & 0.364 & 1.03 & 5.34 \\
    \midrule
    \multicolumn{8}{l}{\emph{(b) PickScore-driven search ($v_\mathrm{sem}\!=\!$\,PickScore)}} \\
    Teacher argmax ($N_t\!=\!8$) & 0.297 & 78.9 & 24.14 & 0.322 & 0.379 & \textbf{1.25} & \textbf{5.35} \\
    \textbf{\textsc{ProxySearch} ($K\!=\!8$)} & \textbf{0.232} & 79.2 & 24.14 & 0.322 & 0.378 & 1.24 & \textbf{5.35} \\
    Teacher-seq ($N_t\!=\!16$, $2\!\times$ NFE) & 0.247 & \textbf{81.2} & \textbf{24.15} & \textbf{0.323} & \textbf{0.384} & 1.19 & \textbf{5.35} \\
    \midrule
    \multicolumn{8}{l}{\emph{(c) HPSv2-driven search ($v_\mathrm{sem}\!=\!$\,HPSv2)}} \\
    Teacher argmax ($N_t\!=\!8$) & 0.296 & 79.0 & 23.91 & \textbf{0.329} & \textbf{0.377} & \textbf{1.27} & \textbf{5.35} \\
    \textbf{\textsc{ProxySearch} ($K\!=\!8$)} & \textbf{0.234} & 78.7 & 23.88 & \textbf{0.329} & \textbf{0.377} & 1.26 & \textbf{5.35} \\
    Teacher-seq ($N_t\!=\!16$, $2\!\times$ NFE) & 0.251 & \textbf{79.2} & \textbf{23.95} & 0.328 & \textbf{0.377} & 1.24 & \textbf{5.35} \\
    \bottomrule
  \end{tabular}%
  }
\end{table}

\smallbreak
\noindent
\textbf{Main result.}
\textsc{ProxySearch} tightens the statistic under every search objective: the paired $\ell_1$ delta is $-0.070\!\pm\!0.004$ SE in panel (a), $-0.065$ in panel (b), and $-0.062$ in panel (c). In panel (a), the paired $t$ is $-19.3$ and the Wilcoxon signed-rank $z$ is $-18.5$. On the optimized reward, \textsc{ProxySearch} and teacher argmax are within noise of each other in every panel, and the held-out rewards behave the same way (App.~\ref{sec:appendix:hero-ci}). The proxy filter does not depend on $v_\mathrm{sem}$, although the final teacher argmax does.

\smallbreak
\noindent
\textbf{Teacher-seq.}
At $2\times$ the teacher NFE, Teacher-seq stays looser than \textsc{ProxySearch} on the statistic in every panel and is ahead on several semantic columns in panels (b) and (c).

\smallbreak
\noindent
\textbf{More teacher candidates do not substitute for the proxy.}
On the first $200$ GenEval prompts, doubling the teacher-argmax pool to $N_t\!=\!16$ raises GE by $1.25$~pp but leaves $\ell_1$ unchanged ($0.298\!\to\!0.296$, App.~\ref{sec:appendix:teacher-scaling}). On a separate $100$-cell pilot, giving teacher argmax the full $64$-candidate pool that \textsc{ProxySearch} screens also leaves $\ell_1$ unchanged ($0.304\!\to\!0.306$). On that pilot, \textsc{ProxySearch} reaches $0.230$ at the teacher cost of $8$ candidates (App.~\ref{sec:appendix:matched-pool}). The gain comes from ranking on the statistic, not from pool size.

\begin{figure}[h]
  \centering
  \includegraphics[width=\linewidth]{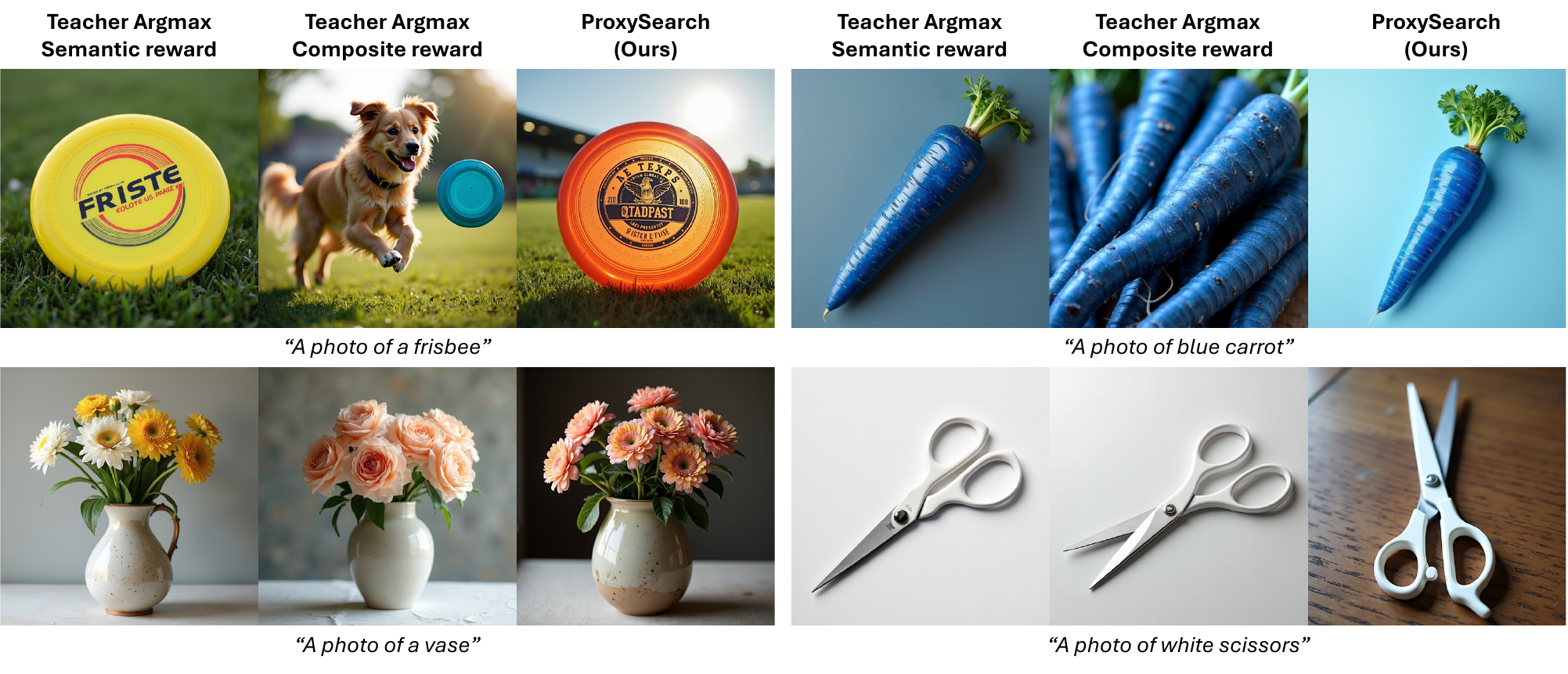}
  \vspace{-6mm}
  \caption{%
    \textbf{Qualitative comparison at fixed noise seed.}
    Four prompts (rows) rendered by teacher argmax, composite reward, and \textsc{ProxySearch} (ours) over the same $32$-noise pool with HPSv2 as the semantic verifier.
    Rows~1--2 target $\tau_{\mathrm{bright}}\!=\!(\mathrm{brightness}\!=\!0.72,\,\mathrm{contrast}\!=\!0.26)$ (bright, high contrast); rows~3--4 target $\tau_{\mathrm{dark}}\!=\!(\mathrm{brightness}\!=\!0.22,\,\mathrm{contrast}\!=\!0.20)$ (dark, low contrast).
    On these four prompts, \textsc{ProxySearch} attains the lowest statistical $\ell_1$ to the target among the three methods while keeping HPSv2 and GenEval comparable; aggregate results are in Tab.~\ref{tab:hero} and Tab.~\ref{tab:method-ablation}.
  }
  \label{fig:qualitative}
  \vspace{-3mm}
\end{figure}

\subsection{Post-processing, composite reward, and human judgments}
\label{sec:exp:alternatives}

Tab.~\ref{tab:method-ablation} compares \textsc{ProxySearch} with post-processing and a composite reward in a smaller $100$-cell pilot at $512^{2}$. Teacher NFE is not matched there: the baselines spend $800$ teacher NFE and \textsc{ProxySearch} $200$.

\begin{table}[t]
  \centering
  \small
  \caption{%
    \textbf{Comparison with alternatives (GE-driven, $100$-cell pilot).} FLUX at $512^{2}$, $50$ GenEval prompts $\times$ $2$ targets, $64$-noise pool. Teacher argmax and composite reward select among $N_t\!=\!16$ teacher renders. \textsc{ProxySearch} keeps $K\!=\!4$ survivors from a $4$-step proxy ($800$ vs.\ $200$ teacher NFE). Random sampling is the teacher render of the first noise. Both post-process rows edit the teacher-argmax image.%
  }
  \label{tab:method-ablation}
  \resizebox{\textwidth}{!}{%
  \begin{tabular}{lccccccc}
    \toprule
    Method & statistical $\ell_1\!\downarrow$ & GE \%$\uparrow$ & PickScore $\uparrow$ & HPSv2 $\uparrow$ & CLIP $\uparrow$ & ImageReward $\uparrow$ & aesthetic $\uparrow$ \\
    \midrule
    Random sampling & 0.303 & 81.0 & 23.11 & 0.302 & 0.354 & 0.96 & \textbf{5.34} \\
    Teacher argmax ($N_t\!=\!16$) & 0.302 & 87.0 & 23.18 & 0.303 & 0.348 & 0.97 & \textbf{5.34} \\
    Composite reward ($N_t\!=\!16$, $\lambda\!=\!0.1$) & 0.205 & 86.5 & \textbf{23.31} & \textbf{0.306} & 0.352 & 1.02 & \textbf{5.34} \\
    Post-process shift (full) & \textbf{0.066} & 84.0 & 22.95 & 0.288 & 0.351 & 0.90 & 5.32 \\
    Post-process shift (clamped to statistical $\!\approx\!0.22$) & 0.218 & \textbf{89.0} & 23.19 & 0.301 & 0.348 & 0.96 & \textbf{5.34} \\
    \textbf{\textsc{ProxySearch} ($K\!=\!4$, $4$-step proxy)} & 0.230 & \textbf{89.0} & 23.26 & \textbf{0.306} & \textbf{0.355} & \textbf{1.04} & \textbf{5.34} \\
    \bottomrule
  \end{tabular}%
  }
  \vspace{-6mm}
\end{table}

\smallbreak
\noindent
\textbf{Post-processing.}
Editing the teacher-argmax image moves the statistic directly. The unclamped shift reaches statistical $\ell_1$ of $0.066$ but trails \textsc{ProxySearch} on GE and on every learned semantic reward. Against the clamped shift, \textsc{ProxySearch} ties on GE ($89.0$), is slightly looser on statistics ($0.230$ vs.\ $0.218$), and is higher on PickScore, HPSv2, CLIP and ImageReward (aesthetic tied). The post-process operations are in App.~\ref{sec:appendix:details}.

\smallbreak
\noindent
\textbf{Composite reward.}
On the $64$-candidate pilot of Sec.~\ref{sec:exp:headline}, a composite reward that scores all $64$ teacher renders reaches a tighter statistic than \textsc{ProxySearch} ($0.143$ vs.\ $0.230$) at $6.9\times$ the total NFE ($8\times$ the teacher NFE). \textsc{ProxySearch} recovers about half of the composite's $\ell_1$ reduction at the teacher cost of $8$ candidates (App.~\ref{sec:appendix:matched-pool}). In the pilot of Tab.~\ref{tab:method-ablation}, the composite reaches $0.205$ and a higher PickScore (HPSv2 tied), while \textsc{ProxySearch} is higher on GE, CLIP, and ImageReward. We sweep $\lambda\!\in\!\{0, 0.1, 0.5, 1, 2\}$ and report $\lambda\!=\!0.1$. Because the GE term is a binary pass and $\ell_1\!<\!1$ within a pool, every $\lambda\!\in\!\{0.1, 0.5, 1, 2\}$ selects the same image on the evaluated cells, and $\lambda\!=\!0$ reduces to teacher argmax. This holds for these runs only; with a much larger $\lambda$, the composite could give up a GE pass for a large $\ell_1$ gain. Fig.~\ref{fig:qualitative} shows example outputs of teacher argmax, the composite reward and \textsc{ProxySearch}.

\smallbreak
\noindent
\textbf{Human judgments.}
In a forced-choice study (App.~\ref{sec:appendix:human-study}), annotators saw the prompt and the requested look on pairs where \textsc{ProxySearch} had shifted the statistic toward that look. They chose the \textsc{ProxySearch} image over the teacher-argmax image in $71.7\%$ of $600$ judgments. This arm measures joint content-plus-look fulfillment. Rates differ by look: $82.0$--$88.4\%$ for dark, warm, and cool, $70.1\%$ for high sharpness, and at chance for bright/high-contrast ($48.1\%$) and low sharpness ($48.2\%$). In a second arm, annotators were told that one image was generated with the look and the other was an ordinary photo edited afterwards, and asked which looks like the genuine unedited photograph. Against the teacher-argmax image post-processed toward the target, with the unclamped shift for bright and dark looks, they chose the \textsc{ProxySearch} image in $84.2\%$ of $400$ judgments. Both images are generated, so this measures perceived naturalness under that framing. For non-tone looks the edit differs from Tab.~\ref{tab:method-ablation} (App.~\ref{sec:appendix:human-study}). When no look is named, a small pilot shows no preference either way ($48.9\%$, $88/180$).

\subsection{Survivor budget and other backbones}
\label{sec:exp:ksweep}

\smallbreak
\noindent
\textbf{Survivor budget.}
We sweep the survivor count $K\!\in\!\{2,4,8,16\}$ on a $100$-cell FLUX pilot ($50$ GenEval prompts $\times$ $2$ targets, $N\!=\!64$ proxy candidates, GenEval as the selection reward), and hold teacher argmax at $N_t\!=\!K$, so both methods run the same number of teacher trajectories except in the $K\!=\!4$ row (Tab.~\ref{tab:ksweep}). Statistical $\ell_1$ tightens by $21.8$--$25.3\%$ at every $K$, while GE stays level with teacher argmax ($+0.5$ to $+2.5$~pp, within noise at this size). The deployed $1$-step proxy at $K\!=\!8$ gives the same picture ($-24.7\%$ $\ell_1$, $+1.0$~pp GE).

\begin{table}[h]
  \centering
  \small
  \caption{%
    \textbf{$K$-sweep at matched teacher trajectories} ($n\!=\!100$ cells per row: $50$ prompts $\times$ $2$ targets, $1024^{2}$). $\Delta$GE: paired difference with $95\%$ CI from a bootstrap over prompts. $^{*}$The $K\!=\!4$ row reuses the run of Tab.~\ref{tab:method-ablation} ($512^{2}$), whose teacher argmax uses $N_t\!=\!16$ rather than $N_t\!=\!K$. Noise pools differ across rows; compare within a row.%
  }
  \vspace{-1mm}
  \resizebox{\linewidth}{!}{%
  \begin{tabular}{cccccccc}
    \toprule
    $K$ & proxy steps & Argmax GE & Ours GE & $\Delta$GE [$95\%$ CI] & Argmax stat.\ $\ell_1$ & Ours stat.\ $\ell_1$ & $\Delta$stat.\,\% \\
    \midrule
    $2$  & $4$ & $81.0$ & $81.5$ & $+0.5\,[-3.0,+4.5]$ & $0.302$ & $0.226$ & $-25.1$ \\
    $4^{*}$ & $4$ & $87.0$ & $89.0$ & $+2.0\,[-2.0,+6.5]$ & $0.302$ & $0.230$ & $-23.8$ \\
    $8$  & $4$ & $87.0$ & $89.5$ & $+2.5\,[-0.5,+6.0]$ & $0.302$ & $0.226$ & $-25.3$ \\
    $16$ & $4$ & $89.0$ & $90.5$ & $+1.5\,[-3.0,+7.5]$ & $0.300$ & $0.235$ & $-21.8$ \\
    \midrule
    $8$  & $1$ & $88.0$ & $89.0$ & $+1.0\,[-3.5,+5.5]$ & $0.304$ & $0.229$ & $-24.7$ \\
    \bottomrule
  \end{tabular}%
  }
  \label{tab:ksweep}
  \vspace{-3.5mm}
\end{table}

\smallbreak
\noindent
\textbf{Other backbones.}
With $\tau_\mathrm{bright}$, $N\!=\!64$, $K\!=\!N_t\!=\!8$, and $k_\mathrm{proxy}\!=\!1$ on $50$ GenEval prompts shared by both backbones ($n\!=\!50$, Tab.~\ref{tab:cross-arch}), \textsc{ProxySearch} reduces statistical $\ell_1$ from $0.204$ to $0.120$ ($41\%$) on SD3.5-Large and from $0.305$ to $0.267$ ($12\%$) on SDXL, while GE drops from $92.0$ to $88.0$ and from $83.0$ to $81.0$.

\begin{table}[h]
  \centering
  \small
  \caption{%
    GenEval and statistical $\ell_1$ to $\tau\!=\!(\text{brightness}\!=\!0.72, \text{contrast}\!=\!0.26)$ at $N\!=\!64$, $K\!=\!N_t\!=\!8$, $k_\mathrm{proxy}\!=\!1$, $n\!=\!50$ GenEval prompts per pair (the same $50$ on both backbones).
    Statistical tightening reproduces on both backbones shown (MMDiT, UNet; DiT in Tab.~\ref{tab:hero} and Tab.~\ref{tab:ksweep}); GenEval is $2$--$4$~pp lower than teacher argmax on both.%
  }
  \label{tab:cross-arch}
  \begin{tabular}{lcccc}
    \toprule
    Pair & Argmax GE & Ours GE & Argmax statistical $\ell_1$ & Ours statistical $\ell_1$ \\
    \midrule
    SD3.5-Large-Turbo~$\to$~SD3.5-Large  & $92.0$ & $88.0$ & $0.204$ & $0.120$ \\
    SDXL-Turbo~$\to$~SDXL                & $83.0$ & $81.0$ & $0.305$ & $0.267$ \\
    \bottomrule
  \end{tabular}
  \vspace{-5mm}
\end{table}
\section{Conclusion}
\label{sec:conclusion}

Major diffusion teachers ship with a distilled few-step companion that inference-time search treats only as a faster generator. We show that this companion agrees with the teacher's noise ranking on closed-form image statistics but not on learned semantic rewards, and that the first denoising step seeds this rank while distillation extends it to finer statistics. \textsc{ProxySearch} uses this split for search under a \emph{joint constraint}: the proxy filters candidates by statistical match, and the teacher selects by semantic reward, with no trade-off weight. At the same number of teacher trajectories, it tightens the statistical target while matching teacher argmax on the optimized reward.
Its benefit is limited to targets on axes the proxy ranks, and we have not yet trained a downstream model on statistic-balanced data. Rank agreement could serve as a distillation objective and as a metric reported next to FID, and the same measurement can be run for video, audio, and 3D distillation. More broadly, the distilled models released alongside each teacher are useful for inference-time scaling, not only for fast sampling.

\clearpage
\begin{ack}
This research was supported by the Scaleup TIPS grant (RS-2023-00321784: Development of Novel Generative AI Technology to Generate Domain-Specific Synthetic Data).
This work was also supported by the National Research Foundation of Korea (NRF) grant funded by the Korea government (MSIT) (RS-2026-25598824) and the Korea Agency for Infrastructure Technology Advancement (KAIA) grant funded by the Ministry of Land, Infrastructure and Transport (RS-2026-25530752).
The authors declare no competing interests.
\end{ack}

\medskip
{\small

\bibliographystyle{ieeenat_fullname}
\bibliography{main}

}

\clearpage

\appendix

\appendix

\section{Experimental details}
\label{sec:appendix:details}

This section lists the models, protocols, compute, verifiers and prompts behind every experiment in the main text.

\subsection{Model pairs and sampling}

Tab.~\ref{tab:protocol} lists the teacher--proxy pairs. The image pairs span three backbones (DiT, MMDiT, UNet). With the DMD2 score-distilled SDXL proxy, which enters only the rank-correlation study, they cover three distillation objectives (LADD, ADD, score). FLUX-dev runs $50$ steps with guidance embedding $3.5$ in a single forward pass (\texttt{true\_cfg\_scale} kept at $1.0$). SD3.5-Large runs $28$ steps and SDXL $50$ steps with classifier-free guidance $4.5$ and $7.5$, using batch-doubled conditional and unconditional passes. The three $1$-step proxies run with guidance $0$. Images are scored at $1024^{2}$ unless noted. The rank-correlation pool of Sec.~\ref{sec:exp:transfer} and its appendix variants, the high-pass control (App.~\ref{sec:appendix:fourier}) and the method-ablation pilot (Tab.~\ref{tab:method-ablation}) run at $512^{2}$.

Experiments run on NVIDIA A100-80GB GPUs with PyTorch $2.10$, CUDA $12.8$ and cuDNN $9$. Each run fits on one GPU. The $553$-prompt headline and the joint-constraint ablations were sharded across several GPUs for wall-clock time. Proxy and teacher transformers run in bf16 and the verifier heads (CLIP, ImageReward, aesthetic) in fp32.

\begin{table}[t]
  \centering
  \small
  \caption{%
    \textbf{Experimental protocol.}
    Teacher--proxy pairs covering DiT, MMDiT, and UNet backbones, plus a DMD2 score-distilled SDXL row used for the rank-correlation extension in App.~\ref{sec:appendix:per-arch}.
    Image pairs share the $11$-verifier bank ($8$ statistics + $3$ semantics).
    $^{\ddagger}$$k_\mathrm{proxy}\!=\!1$ is the operational setting used by \textsc{ProxySearch}; raising the proxy to
    $k_\mathrm{proxy}\!=\!4$ leaves brightness and color-temperature rank unchanged and raises some weaker axes by up to $0.11$ on FLUX and SD3.5 (App.~\ref{sec:appendix:proxy-step-count}), and lowers most axes on SDXL (Tab.~\ref{tab:appendix:per-arch-image}); the method ablation (Tab.~\ref{tab:method-ablation}) and the $K$-sweep of Tab.~\ref{tab:ksweep} use a $4$-step proxy.%
  }
  \begin{tabular}{lllrrr}
    \toprule
    Pair & Backbone & Distillation & $k_\mathrm{proxy}$ & teacher steps & verifiers \\
    \midrule
    FLUX-schnell~$\to$~FLUX-dev          & DiT   & LADD & $1^{\ddagger}$ & 50 & 8 stat + 3 sem \\
    SD3.5-Large-Turbo~$\to$~SD3.5-Large  & MMDiT & ADD  & $1^{\ddagger}$ & 28 & 8 stat + 3 sem \\
    SDXL-Turbo~$\to$~SDXL                & UNet  & ADD  & $1$ & 50 & 8 stat + 3 sem \\
    SDXL+DMD2~$\to$~SDXL                  & UNet  & score        & $1$ & 50 & 8 stat + 3 sem \\
    \bottomrule
  \end{tabular}
  \label{tab:protocol}
\end{table}

\subsection{Per-experiment protocol}

Tab.~\ref{tab:appendix:exp-protocol} gives one row per experiment. Absolute GE is not comparable across tables because the prompt sets differ. On FLUX, teacher argmax reads $81.6\%$ in Tab.~\ref{tab:hero} ($553$ GenEval prompts, $N_t\!=\!8$), $93.75\%$ in Tab.~\ref{tab:appendix:nt-scaling} (first $200$ prompts in file order, only single-object, two-object and counting, $N_t\!=\!8$), $87.0\%$ in Tab.~\ref{tab:method-ablation} ($N_t\!=\!16$, $512^{2}$) and $81.0$--$89.0\%$ in Tab.~\ref{tab:ksweep} ($N_t\!=\!K$). GE also varies strongly by category on the headline run (teacher argmax: single-object $96.3\%$, position $58.5\%$). Comparisons are meaningful within a table, where methods share prompts, targets and the noise pool of one run.

Noises come from a helper \texttt{noise\_from\_id(base\_seed, prompt\_id, candidate\_idx)} (base seed $42$ for search runs) that returns the same noise for the same triple regardless of batch size. Within a run, proxy and teacher renders of candidate $i$ use the same tensor, so every within-run comparison is paired. The helper's seed depends on Python's string hash, so a pool is reproducible across processes only when \texttt{PYTHONHASHSEED} is fixed. Several runs were launched without a fixed hash seed, and their pools cannot be regenerated: the rank-correlation runs (Tab.~\ref{tab:rho} and its appendix variants), the $N_t\!=\!16$ arm of Tab.~\ref{tab:appendix:nt-scaling}, the run behind Tab.~\ref{tab:method-ablation}, the $K\!=\!2,8,16$ $K$-sweep runs and the three RL arms. Comparisons across these runs are paired by prompt and target, not by noise. The null controls (App.~\ref{sec:appendix:null-controls}) use a saved noise tensor.

\begin{table}[t]
  \centering
  \footnotesize
  \caption{%
    \textbf{Per-experiment protocol.} Teacher steps as in Tab.~\ref{tab:protocol} unless listed. \emph{Cells} are prompt--target pairs (prompts for rank-correlation studies, which use no $\tau$). $N$: noises per cell (proxy candidates in search experiments). $K$: teacher survivors. Teacher argmax uses $N_t\!=\!K$ unless noted. In search experiments the proxy retains the top-$K$ by $g_\mathrm{stat}$ and the teacher argmaxes the listed selection reward. \emph{Template}: $100$ synthetic template prompts (App.~\ref{sec:appendix:prompts}). \emph{pilot50}: $50$ category-stratified GenEval prompts. n/r: not recorded.%
  }
  \label{tab:appendix:exp-protocol}
  \setlength{\tabcolsep}{4pt}
  \begin{tabular}{@{}>{\raggedright\arraybackslash}p{2.9cm}>{\raggedright\arraybackslash}p{4.3cm}>{\raggedright\arraybackslash}p{3.5cm}>{\raggedright\arraybackslash}p{2.3cm}@{}}
    \toprule
    Experiment & Pair(s); resolution; prompts; cells & Sampling and selection & Uncertainty \\
    \midrule
    \multicolumn{4}{@{}l}{\emph{Rank-correlation studies (no $\tau$, no selection; Spearman $\rho$)}} \\
    Rank transfer (Tab.~\ref{tab:rho}) & FLUX, SD3.5, SDXL; $512^{2}$; $100$ template; $100$/pair & $N\!=\!32$; $k_\mathrm{proxy}\!=\!1$ & $95\%$ Wald CI; $\sigma$; sig\% \\
    Score distillation (Tab.~\ref{tab:rho}, DMD2) & SDXL+DMD2; $512^{2}$; $100$ template; $100$ & $N\!=\!32$; $k_\mathrm{proxy}\!=\!1$ & $\sigma$; sig\% \\
    Proxy step count (App.~\ref{sec:appendix:proxy-step-count}) & FLUX, SD3.5 (SDXL: rank-transfer pool); $512^{2}$; $100$ template; $100$/pair & $N\!=\!32$; $k_\mathrm{proxy}\!\in\!\{1,4\}$ & none printed \\
    Proxy-semantic top-$1$ (Tab.~\ref{tab:appendix:proxy-hf-ceiling}) & FLUX, SD3.5; $512^{2}$; $100$ template; $100$/pair & $N\!\in\!\{1,2,4,$ $8,16,32\}$; $k_\mathrm{proxy}\!=\!4$ (FLUX), n/r (SD3.5); proxy top-$1$ by CLIP, ImageReward, aesthetic & none (prompt mean) \\
    High-pass control (App.~\ref{sec:appendix:fourier}) & FLUX; $512^{2}$; $50$ hand-written; $50$ & $N\!=\!32$; $k_\mathrm{proxy}\!=\!1$ & none \\
    Paired null controls (Tab.~\ref{tab:appendix:paired-null}) & FLUX, SD3.5; $1024^{2}$; $12$ GenEval (stratified); $12$/pair & $N\!=\!32$ (shared); $k_\mathrm{proxy}\!=\!1$ ($4$ as reference); teacher $50$/$28$ steps (reference), $1$ step (null) & $95\%$ noise-bootstrap CI; permutation $p$ \\
    Shared-noise / learned predictors (App.~\ref{sec:appendix:shared-noise}) & FLUX, SD3.5; $1024^{2}$; $12$ GenEval (stratified); $12$/pair & $N\!=\!32$ (shared); $k_\mathrm{proxy}\!=\!4$ & noise-bootstrap CI; permutation $p$ \\
    RL post-training (App.~\ref{sec:appendix:rl-postrain}) & SD3.5-Medium proxy~$\to$~SD3.5-Medium (frozen / text / PickScore LoRA); $1024^{2}$; $100$ FlowGRPO ($50$ OCR test + $50$ PickScore); $100$ & $N\!=\!32$; $k_\mathrm{proxy}\!=\!1$; teacher $28$ steps & $95\%$ CI half-width over prompts \\
    \midrule
    \multicolumn{4}{@{}l}{\emph{Search experiments (proxy top-$K$ by $g_\mathrm{stat}$, teacher argmax of the selection reward)}} \\
    Headline (Tab.~\ref{tab:hero}) & FLUX; $1024^{2}$; $553$ GenEval $\times$ $2$ $\tau$; $1106$ & $N\!=\!64$; $K\!=\!8$ (Teacher-seq: $16\!\to\!8$, $100$ cells measured); $k_\mathrm{proxy}\!=\!1$; GE / PickScore / HPSv2 & paired bootstrap $95\%$ CI; paired $t$ (a) \\
    Method ablation (Tab.~\ref{tab:method-ablation}) & FLUX; $512^{2}$; pilot50 $\times$ $2$ $\tau$; $100$ & $N\!=\!64$; $K\!=\!4$ ($N_t\!=\!16$); $k_\mathrm{proxy}\!=\!4$; GE & none \\
    $K$-sweep (Tab.~\ref{tab:ksweep}) & FLUX; $1024^{2}$ ($K\!=\!4$: $512^{2}$); pilot50 $\times$ $2$ $\tau$; $100$ & $N\!=\!64$; $K\!\in\!\{2,4,8,16\}$ ($N_t\!=\!K$; $K\!=\!4$: $N_t\!=\!16$); $k_\mathrm{proxy}\!=\!4$ ($1$ for one $K\!=\!8$ row); GE & paired prompt-bootstrap $95\%$ CI \\
    Matched pool (App.~\ref{sec:appendix:matched-pool}) & FLUX; $1024^{2}$; pilot50 $\times$ $2$ $\tau$; $100$ & $N\!=\!64$; $K\!=\!8$ ($N_t\!\in\!\{8,64\}$); $k_\mathrm{proxy}\!=\!1$; GE & prompt-cluster bootstrap $95\%$ CI \\
    Cross-backbone (Tab.~\ref{tab:cross-arch}) & SD3.5-Large, SDXL; $1024^{2}$; $50$ GenEval (stratified, same for both) $\times$ $1$ $\tau$; $50$/pair & $N\!=\!64$; $K\!=\!8$; $k_\mathrm{proxy}\!=\!1$; GE & none \\
    Teacher-pool scaling (Tab.~\ref{tab:appendix:nt-scaling}) & FLUX; $1024^{2}$; first $200$ GenEval $\times$ $2$ $\tau$; $400$ & $N\!=\!64$; $K\!=\!8$ ($N_t\!\in\!\{8,16\}$); $k_\mathrm{proxy}\!=\!1$; GE & none \\
    User study (App.~\ref{sec:appendix:human-study}) & FLUX; $1024^{2}$ (shown $\le\!512$ px); GenEval $\times$ $6$ looks; $144$ / $97$ / $180$ pairs & $N\!=\!64$; $K\!=\!8$; $k_\mathrm{proxy}\!=\!1$; GE; $21$ items from the Tab.~\ref{tab:method-ablation} pilot ($N_t\!=\!16$, $K\!=\!4$, $4$-step proxy) & Wilson; two-way bootstrap \\
    \bottomrule
  \end{tabular}
\end{table}

\subsection{Compute accounting}
\label{sec:appendix:compute}

NFE counts denoising steps only (Sec.~\ref{sec:prelim}). It excludes classifier-free-guidance passes, VAE decoding, text encoding and verifier scoring. With a $1$-step proxy, $c_\mathrm{proxy}/c_\mathrm{teacher}$ is $1/50$ on FLUX-$1024$ and SDXL and $1/28$ on SD3.5. Counting the teacher's guidance passes, it is $1/56$ on SD3.5 and $1/100$ on SDXL. At $N\!=\!64$, $K\!=\!N_t\!=\!8$, teacher argmax and \textsc{ProxySearch} then cost $448$ and $512$ passes on SD3.5 and $800$ and $864$ on SDXL. FLUX-dev uses a single guidance-embedded pass, so its counts equal the NFE ($400$ and $464$).

Measured render time per cell grows more than the NFE ratio, because each proxy render carries its own VAE decode and text encoding: about $+24\%$ on FLUX ($186$ vs.\ $230$~s), $+28\%$ on SD3.5 and $+49\%$ on SDXL. These timings exclude scoring. They are approximate, since the teacher-only time is apportioned from a shared teacher pool rather than timed separately.

\subsection{Verifiers and selection rules}

\paragraph{Statistics.} All eight statistics are computed with numpy and PIL on the decoded image at its native resolution, without normalization: brightness (mean RGB$/255$), contrast (std of grayscale$/255$), saturation (mean HSV saturation$/255$), colorfulness (Hasler--S\"usstrunk), sharpness (variance of the $3\!\times\!3$ Laplacian on grayscale), edge density (fraction of pixels whose Sobel magnitude exceeds its mean plus one std), color temperature ($\log$ of mean R over mean B) and compressibility (JPEG quality-$95$ size in KB, so a larger value means a less compressible image).

\paragraph{Semantic verifiers.} The rank-transfer bank uses CLIP score~\citep{radford2021clip,cherti2023openclip} (\texttt{open\_clip} ViT-H-14), ImageReward~\citep{xu2023imagereward} (\texttt{ImageReward-v1.0}) and the aesthetic predictor~\citep{schuhmann2022aesthetic,schuhmann2022laion5b} (LAION v1, CLIP ViT-L/14 backbone with a $768\!\to\!1024\!\to\!128\!\to\!64\!\to\!16\!\to\!1$ MLP head). The PickScore- and HPSv2-driven searches (Tab.~\ref{tab:hero}, panels b and c) replace the GE search objective with PickScore (\texttt{PickScore\_v1}, CLIP-H dual encoder) or HPSv2 (\texttt{hpsv2} package, v2.1) on the same $n\!=\!1106$ paired cells.

\paragraph{Statistical objective.} $g_\mathrm{stat}$ sums the absolute deviations over the axes named in $\tau$ with unit weights. $\tau_\mathrm{bright}\!=\!(\mathrm{brightness}\!=\!0.72,\,\mathrm{contrast}\!=\!0.26)$ and $\tau_\mathrm{dark}\!=\!(0.22,\,0.20)$ are fixed presets shared by all prompts. The proxy top-$K$ is taken by sorting the continuous proxy $\ell_1$. $g_\mathrm{stat}$ can also encode axis-pushing ($\pm\ell_j$ for the $j$-th axis) or thresholding ($\mathbf{1}[\ell_j\!\ge\!t]$), and the algorithm applies to any $g_\mathrm{stat}$ for which larger values are preferred. All reported experiments use the target-matching form. Indicator-valued objectives can tie at the top-$K$ cut and then need a tie-breaking rule.

\paragraph{Selection rules.} GE-driven selection (teacher argmax and the \textsc{ProxySearch} teacher stage alike) takes the lexicographic maximum of (in-process GenEval pass, detector confidence, smaller teacher-image $\ell_1$). The confidence is the maximum detection score over all detected classes. On the stored teacher pools we checked, the first two keys were never tied, so $\ell_1$ did not decide a pick. PickScore- and HPSv2-driven selection is a plain argmax. Ties go to the first candidate: the lowest noise index for teacher argmax and the smallest proxy $\ell_1$ for \textsc{ProxySearch}. HPSv2 scores tie at the maximum in $18$ of the $1106$ cells for teacher argmax and in $17$ for \textsc{ProxySearch}.

\paragraph{Post-process baselines.} The unclamped shift (Tab.~\ref{tab:method-ablation}) edits the teacher-argmax image. It adds $\tau_\mathrm{brightness}\!-\!\mathrm{brightness}$ to every RGB channel (in $[0,1]$, then clipped) and applies a PIL contrast enhancement with factor $\tau_\mathrm{contrast}/\mathrm{contrast}$ clipped to $[0.2, 3]$. The clamped variant is the pixel blend $0.68\,x_\mathrm{argmax}\!+\!0.32\,x_\mathrm{shift}$ with one fixed weight for all cells (resulting statistical $\ell_1$ $0.218$). Neither costs NFE.

\subsection{Prompt banks}
\label{sec:appendix:prompts}

\paragraph{GenEval.} The headline uses the full $553$-prompt GenEval evaluation set $\times$ $2$ targets ($1106$ paired cells). The same cells feed all three panels of Tab.~\ref{tab:hero}. The $100$-cell FLUX pilots (Tabs.~\ref{tab:method-ablation}, \ref{tab:ksweep} and App.~\ref{sec:appendix:matched-pool}) use pilot50, $50$ GenEval prompts sampled with stratification by category (seed $42$), $\times$ $2$ targets. Tab.~\ref{tab:cross-arch} uses a second stratified $50$-prompt sample ($41$ prompts shared with pilot50) and one target. The Teacher-seq measurement uses a third $50$-prompt GenEval sample.

\paragraph{Rank-correlation template.} The rank-transfer pool of Sec.~\ref{sec:exp:transfer} and its appendix variants use $100$ synthetic template prompts $\times$ $32$ noises at $512^{2}$. Five templates are filled from small vocabularies of colors, objects, materials, locations, actions and counts (seed $42$), for example ``a bird colored 3 with book 2''. None is a GenEval prompt, and the same $100$ prompts are used for every pair.

\paragraph{FlowGRPO.} The RL study (App.~\ref{sec:appendix:rl-postrain}) uses $100$ FlowGRPO prompts (\href{https://github.com/yifan123/flow_grpo}{\texttt{github.com/yifan123/flow\_grpo}}): the first $50$ prompts of its OCR test split, for example ``A close-up of a sleek smartwatch on a wrist, the screen displaying \emph{Step Goal Achieved} with a celebratory animation, [\ldots]'', and $50$ PickScore prompts, for example ``gold tip pyramid in the night, extremely detailed, rain, stars''.

\section{Rank transfer: full results}
\label{sec:appendix:per-arch}

This section gives the numbers behind Fig.~\ref{fig:transfer}~(a) and the robustness checks of the statistics/semantics split (Sec.~\ref{sec:exp:transfer} and Sec.~\ref{sec:exp:rl-postrain}).

\subsection{Per-verifier rank correlation}

\begin{table}[t]
  \centering
  \footnotesize
  \caption{%
    \textbf{Per-verifier proxy--teacher Spearman $\rho$.} $n\!=\!100$ template prompts $\times$ $32$ noises at $512^{2}$, $1$-step proxy vs.\ full teacher ($50$ steps for FLUX and SDXL, $28$ for SD3.5). $\rho$: mean per-prompt Spearman. $\sigma$: its standard deviation across prompts. sig\%: proportion of prompts at $p\!<\!0.05$ on the within-prompt Spearman test (chance is $5\%$). The $95\%$ Wald CI on each FLUX, SD3.5 and SDXL-Turbo mean has half-width $0.03$--$0.04$. DMD2: the SDXL-Base UNet swapped with the official $1$-step score-distilled DMD2 weights~\citep{yin2024dmd2} (\texttt{dmd2\_sdxl\_1step\_unet\_fp16.bin}), run at $1$ step with guidance scale $0.0$. Within each pair every statistical verifier transfers more strongly than every semantic verifier. Across pairs the ranges overlap at the low end (SDXL-Turbo contrast $0.095$ vs.\ FLUX ImageReward $0.116$).%
  }
  \label{tab:rho}
  \label{tab:appendix:per-arch-image}
  \label{tab:appendix:per-arch-image-distill}
  \setlength{\tabcolsep}{4pt}
  \begin{tabular}{l ccc ccc ccc ccc}
    \toprule
    & \multicolumn{3}{c}{\textbf{FLUX} (LADD)} & \multicolumn{3}{c}{\textbf{SD3.5} (ADD)} & \multicolumn{3}{c}{\textbf{SDXL-Turbo} (ADD)} & \multicolumn{3}{c}{\textbf{SDXL-DMD2} (score)} \\
    \cmidrule(lr){2-4} \cmidrule(lr){5-7} \cmidrule(lr){8-10} \cmidrule(lr){11-13}
    Verifier & $\rho$ & $\sigma$ & sig\% & $\rho$ & $\sigma$ & sig\% & $\rho$ & $\sigma$ & sig\% & $\rho$ & $\sigma$ & sig\% \\
    \midrule
    \multicolumn{13}{l}{\emph{statistics (8 verifiers)}} \\
    brightness         & $0.557$ & $0.227$ & $83$ & $0.804$ & $0.168$ & $99$ & $0.337$ & $0.216$ & $53$ & $0.553$ & $0.131$ & $96$ \\
    color\_temperature & $0.486$ & $0.201$ & $81$ & $0.584$ & $0.175$ & $91$ & $0.336$ & $0.195$ & $51$ & $0.306$ & $0.174$ & $42$ \\
    compressibility    & $0.421$ & $0.193$ & $67$ & $0.567$ & $0.196$ & $88$ & $0.310$ & $0.181$ & $45$ & $0.239$ & $0.196$ & $30$ \\
    saturation         & $0.380$ & $0.211$ & $57$ & $0.538$ & $0.176$ & $82$ & $0.211$ & $0.203$ & $24$ & $0.086$ & $0.195$ & $9$  \\
    sharpness          & $0.322$ & $0.198$ & $46$ & $0.455$ & $0.201$ & $72$ & $0.234$ & $0.174$ & $24$ & $0.190$ & $0.172$ & $19$ \\
    contrast           & $0.314$ & $0.191$ & $47$ & $0.410$ & $0.194$ & $62$ & $0.095$ & $0.197$ & $11$ & $0.159$ & $0.164$ & $14$ \\
    colorfulness       & $0.303$ & $0.198$ & $37$ & $0.263$ & $0.228$ & $39$ & $0.196$ & $0.193$ & $23$ & $0.088$ & $0.170$ & $9$  \\
    edge\_density      & $0.183$ & $0.197$ & $20$ & $0.210$ & $0.201$ & $27$ & $0.145$ & $0.204$ & $17$ & $0.079$ & $0.199$ & $11$ \\
    \midrule
    \multicolumn{13}{l}{\emph{semantics (3 verifiers)}} \\
    CLIP               & $0.095$ & $0.185$ & $12$ & $0.085$ & $0.204$ & $12$ & $0.021$ & $0.170$ & $3$  & $0.022$ & $0.185$ & $6$ \\
    ImageReward        & $0.116$ & $0.181$ & $13$ & $0.114$ & $0.204$ & $16$ & $0.016$ & $0.162$ & $4$  & $-0.006$ & $0.163$ & $4$ \\
    aesthetic          & $0.118$ & $0.188$ & $13$ & $0.069$ & $0.179$ & $6$  & $0.025$ & $0.175$ & $2$  & $0.013$ & $0.172$ & $4$ \\
    \bottomrule
  \end{tabular}
\end{table}

Tab.~\ref{tab:rho} reports the rank correlations for all four pairs. Three patterns hold for every pair: statistical $\rho$ is positive on every verifier, semantic $\rho$ is at most $0.118$, and brightness is the strongest statistical axis. SDXL-Turbo carries the weakest statistical signal of the ADD/LADD pairs (best sig\% $53$ for brightness, vs.\ $99$ on SD3.5). It differs from the other proxies in both backbone (UNet) and distillation recipe, so we do not attribute this to either. Its statistics/semantics separation is still the largest ($\approx\!16\times$: brightness $0.337$ vs.\ CLIP $0.021$).

The DMD2 score-distilled proxy shows the same within-pair split as the LADD and ADD proxies (brightness $0.553$ vs.\ CLIP $0.022$), but not the statistical profile of SDXL-Turbo. Statistical rank transfers in rough magnitude order, from brightness (strongest) to edge density ($0.079$, weakest), which transfers only marginally.

\subsection{Proxy-side semantic selection}

This subsection tests whether the proxy can select on semantics despite the low semantic $\rho$ of Tab.~\ref{tab:rho}. In a separate run on the same $100$ template prompts $\times$ $32$ noises, for each $N\!\in\!\{1,2,4,8,16,32\}$ we take the first $N$ candidates and record the teacher-semantic score of the candidate the proxy ranks first (\emph{proxy}). We compare it with the maximum teacher-semantic score over the same $N$ (\emph{oracle}), the value an unrestricted teacher search would attain. On CLIP and aesthetic the proxy pick stays close to its $N\!=\!1$ value. ImageReward proxy pick rises with $N$ (FLUX $0.239\!\to\!0.460$, SD3.5 $-0.005\!\to\!0.338$) but stays well below the oracle, which grows monotonically with $N$. Semantic gains in this setting come from teacher rollouts.

\begin{table}[t]
  \centering
  \footnotesize
  \caption{%
    \textbf{Proxy-semantic top-1 selection vs.\ the teacher-semantic oracle on the same $N$ candidates.} Prompt-mean teacher-semantic score. $100$ template prompts $\times$ $32$ noises at $512^{2}$. Proxy: $4$ steps on FLUX, step count not recorded on SD3.5. Teacher: $50$ steps (FLUX), $28$ steps (SD3.5). At $N\!=\!32$ the oracle-minus-proxy gap on FLUX is $+0.058$ CLIP, $+0.077$ aesthetic and $+0.773$ ImageReward. SD3.5 follows the same pattern.%
  }
  \label{tab:appendix:proxy-hf-ceiling}
  \begin{tabular}{ll cc cc cc}
    \toprule
    & & \multicolumn{2}{c}{$N\!=\!1$} & \multicolumn{2}{c}{$N\!=\!8$} & \multicolumn{2}{c}{$N\!=\!32$} \\
    \cmidrule(lr){3-4} \cmidrule(lr){5-6} \cmidrule(lr){7-8}
    Pair & semantic verifier & proxy & oracle & proxy & oracle & proxy & oracle \\
    \midrule
    FLUX  & CLIP        & $0.294$ & $0.294$ & $0.294$ & $0.335$ & $0.298$ & $0.356$ \\
    FLUX  & ImageReward & $0.239$ & $0.239$ & $0.318$ & $0.926$ & $0.460$ & $1.233$ \\
    FLUX  & aesthetic   & $5.377$ & $5.377$ & $5.378$ & $5.420$ & $5.376$ & $5.452$ \\
    \midrule
    SD3.5 & CLIP        & $0.334$ & $0.334$ & $0.343$ & $0.380$ & $0.346$ & $0.397$ \\
    SD3.5 & ImageReward & $-0.005$ & $-0.005$ & $0.252$ & $0.933$ & $0.338$ & $1.194$ \\
    SD3.5 & aesthetic   & $5.275$ & $5.275$ & $5.284$ & $5.342$ & $5.299$ & $5.373$ \\
    \bottomrule
  \end{tabular}
\end{table}

\subsection{Proxy step count}
\label{sec:appendix:proxy-step-count}

\begin{figure}[t]
  \centering
  \includegraphics[width=\linewidth]{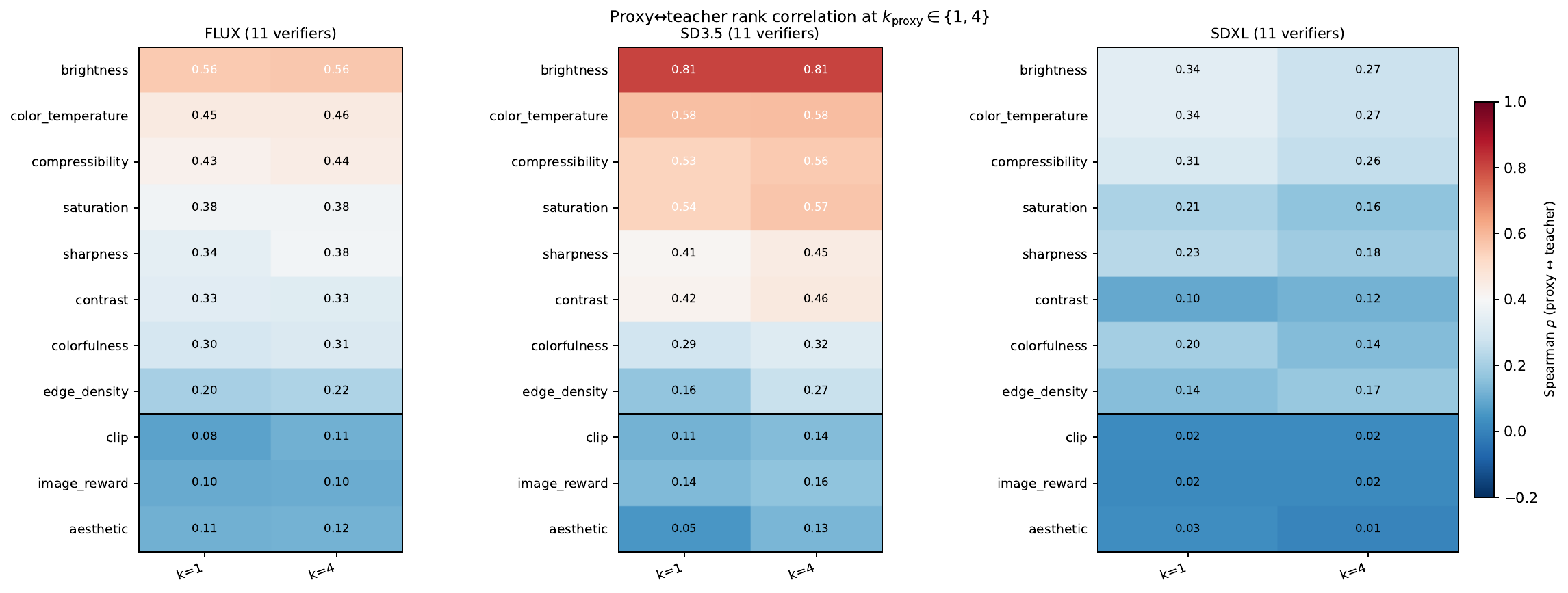}
  \caption{%
    \textbf{A $4$-step proxy does not recover semantic rank.}
    Per-verifier Spearman $\rho$ between the proxy at $k_\mathrm{proxy}\!\in\!\{1,4\}$ and the full teacher, $100$ template prompts $\times$ $32$ noises at $512^{2}$. Panels: FLUX-schnell$\!\to\!$dev, SD3.5-Large-Turbo$\!\to\!$Large and SDXL-Turbo$\!\to\!$SDXL. Rows: eight statistics (top) and three semantic verifiers (bottom). The FLUX and SD3.5 panels come from a separate run with its own noises. The SDXL panel uses the pool of Tab.~\ref{tab:rho}. Semantic $\rho$ stays below $0.16$ at $k_\mathrm{proxy}\!=\!4$ on every verifier and pair.%
  }
  \label{fig:appendix:4step-transfer}
\end{figure}

We test whether more proxy steps recover semantic rank transfer. For FLUX-schnell$\!\to\!$dev and SD3.5-Large-Turbo$\!\to\!$Large we re-run the rank-correlation protocol at $k_\mathrm{proxy}\!\in\!\{1,4\}$ ($100$ template prompts $\times$ $32$ noises at $512^{2}$, all $11$ verifiers). This run draws its own noises (App.~\ref{sec:appendix:details}), so its $k_\mathrm{proxy}\!=\!1$ values differ slightly from Tab.~\ref{tab:rho}, for example SD3.5 ImageReward $0.136$ here vs.\ $0.114$ there. A few values exceed the $0.80$ and $0.12$ bounds quoted for that pool. The SDXL panel of Fig.~\ref{fig:appendix:4step-transfer} uses the pool of Tab.~\ref{tab:rho}.

On FLUX and SD3.5, brightness and color temperature are flat from $k_\mathrm{proxy}\!=\!1$ to $4$ (FLUX brightness $0.559\!\to\!0.559$, SD3.5 brightness $0.807\!\to\!0.807$). Several weaker statistical axes gain significantly at $k_\mathrm{proxy}\!=\!4$ (paired Wilcoxon over prompts): SD3.5 edge density $+0.106$, SD3.5 contrast, colorfulness and sharpness about $+0.04$, and FLUX sharpness $+0.04$. Semantic $\rho$ stays below $0.16$ on every verifier (FLUX CLIP $0.078\!\to\!0.111$, ImageReward $0.099\!\to\!0.104$, aesthetic $0.113\!\to\!0.118$; SD3.5 CLIP $0.114\!\to\!0.138$, ImageReward $0.136\!\to\!0.159$, aesthetic $0.052\!\to\!0.126$). The SD3.5 aesthetic ($p\!=\!0.002$) and FLUX CLIP ($p\!=\!0.046$) gains are nominally significant. The $4$-step proxy is not a usable semantic ranker.

On SDXL, a $4$-step SDXL-Turbo proxy lowers most statistical axes (brightness $0.337\!\to\!0.272$, color temperature $0.336\!\to\!0.270$). Contrast and edge density rise slightly ($0.095\!\to\!0.118$, $0.145\!\to\!0.171$), and the semantic verifiers stay near zero (CLIP $0.021$, aesthetic $0.006$ at $k_\mathrm{proxy}\!=\!4$).

The $1$-step and $4$-step proxy renders rank candidates almost identically on brightness ($\rho\!\ge\!0.99$ on FLUX and SD3.5), less closely on the other statistics (down to $\rho\!=\!0.42$), and less closely still on semantics ($\rho\!\in\![0.22, 0.70]$). A $4$-step render costs about $3\times$ a $1$-step render ($2.05$ vs.\ $0.69$~s on FLUX-$1024$). The deployed $1$-step proxy thus gives up $0.02$--$0.11$ in $\rho$ on several weaker axes for a third of the proxy cost. This is a rank-level comparison. The $K$-sweep (Tab.~\ref{tab:ksweep}) reports one $1$-step search row on a separate pool.

\subsection{Teacher RL post-training}
\label{sec:appendix:rl-postrain}

This subsection gives the protocol and per-cell numbers behind the RL result of Sec.~\ref{sec:exp:rl-postrain}. The proxy is held fixed across conditions. It is the third-party distilled checkpoint \texttt{tensorart/stable-diffusion-3.5-medium-turbo}, run at $1$ step with guidance $1.5$. Its model card recommends $8$ steps and does not state the distillation method. The teacher is SD3.5-Medium ($28$ steps, guidance $4.5$) in three conditions: frozen, with the FlowGRPO text-rendering LoRA (\texttt{jieliu/SD3.5M-FlowGRPO-Text}), and with the FlowGRPO PickScore-reward LoRA (\texttt{jieliu/SD3.5M-FlowGRPO-PickScore})~\citep{liu2025flowgrpo}. Each condition uses $n\!=\!100$ FlowGRPO prompts $\times$ $32$ noises at $1024^{2}$ (App.~\ref{sec:appendix:prompts}). Of the $50$ PickScore prompts, $48$ are in the PickScore LoRA's training split, so that arm is evaluated largely on prompts its teacher was tuned on. Each arm draws its own noises, so arms are paired by prompt only. Fig.~\ref{fig:rl-postrain} shows the per-verifier $\rho$ under each teacher (top) and percentile traversals along three statistical axes (bottom). Tab.~\ref{tab:appendix:rl-postrain} reports per-cell $95\%$ CI half-widths, $1.96\,\sigma/\sqrt{n}$ across the $100$ prompts with $\sigma$ the sample standard deviation.

\begin{table}[ht]
  \centering
  \small
  \caption{%
    \textbf{Proxy$\to$teacher Spearman $\rho$ under three teacher conditions} ($\rho\pm95\%$ CI half-width). The text-rendering LoRA stays within about $0.06$ of the frozen teacher on every statistical axis. The PickScore-reward LoRA lowers statistical $\rho$ by roughly $0.14$--$0.23$ on every axis except edge density ($0.133\!\to\!0.068$), and brightness ($0.684$) still exceeds the best semantic verifier (CLIP, $0.045$) by about $15\times$.%
  }
  \label{tab:appendix:rl-postrain}
  \begin{tabular}{lccc}
    \toprule
    Verifier & Frozen teacher & Text-rendering LoRA & PickScore-reward LoRA \\
    \midrule
    \multicolumn{4}{l}{\emph{statistics (8 verifiers)}} \\
    brightness          & $+0.823 \pm 0.039$ & $+0.829 \pm 0.035$ & $+0.684 \pm 0.052$ \\
    color\_temperature  & $+0.628 \pm 0.050$ & $+0.615 \pm 0.049$ & $+0.460 \pm 0.052$ \\
    compressibility     & $+0.570 \pm 0.044$ & $+0.509 \pm 0.050$ & $+0.347 \pm 0.048$ \\
    saturation          & $+0.491 \pm 0.048$ & $+0.458 \pm 0.048$ & $+0.271 \pm 0.050$ \\
    sharpness           & $+0.400 \pm 0.051$ & $+0.350 \pm 0.052$ & $+0.207 \pm 0.044$ \\
    contrast            & $+0.496 \pm 0.056$ & $+0.508 \pm 0.047$ & $+0.273 \pm 0.054$ \\
    colorfulness        & $+0.432 \pm 0.046$ & $+0.396 \pm 0.047$ & $+0.203 \pm 0.050$ \\
    edge density        & $+0.133 \pm 0.051$ & $+0.113 \pm 0.055$ & $+0.068 \pm 0.040$ \\
    \midrule
    \multicolumn{4}{l}{\emph{semantics (3 verifiers)}} \\
    CLIP score          & $-0.010 \pm 0.036$ & $+0.025 \pm 0.037$ & $+0.045 \pm 0.041$ \\
    ImageReward         & $+0.001 \pm 0.041$ & $-0.009 \pm 0.039$ & $-0.003 \pm 0.033$ \\
    aesthetic           & $+0.072 \pm 0.038$ & $+0.043 \pm 0.042$ & $+0.020 \pm 0.039$ \\
    \bottomrule
  \end{tabular}
\end{table}

\section{Where statistical rank comes from: null controls}
\label{sec:appendix:null-controls}

This section separates what the raw noise, one teacher step and distillation each contribute to the statistical rank (Fig.~\ref{fig:transfer}~(b), Sec.~\ref{sec:exp:transfer}). All controls use one saved pool of $32$ noise tensors and $12$ GenEval prompts stratified by category, at $1024^{2}$. Noise $j$ is the identical $z_T$ for every model and condition. Because all prompts share the $32$ noises, the noise is the independent unit: CIs resample noise indices (the same resample for every prompt), and two-sided permutation tests permute them ($B\!=\!2000$ each).

\subsection{Paired null controls}

Tab.~\ref{tab:appendix:paired-null} compares three $1$-step conditions against the full teacher: the global mean of the raw latent, the undistilled teacher truncated to one step (teacher guidance), and the distilled proxy. The $4$-step proxy is listed for reference. Re-rendering the full teacher for two prompts ($64$ images per backbone) matches the stored teacher statistics at Spearman $0.96$--$1.00$ on every statistic.

The raw-latent mean shows no positive transfer. On FLUX it is weakly anti-correlated with color temperature and saturation. The $1$-step teacher ranks the coarse statistics as well as or better than the $1$-step proxy (brightness, contrast and saturation on both backbones). Distillation contributes on the finer statistics. On FLUX the proxy roughly doubles the $1$-step teacher's $\rho$ on color temperature, compressibility and sharpness, and it is significant on edge density where the teacher is not. On SD3.5 the gain is confined to edge density. The distilled proxy is the only $1$-step condition significant on all eight statistics, and its $1$-step and $4$-step variants agree closely (App.~\ref{sec:appendix:proxy-step-count}).

The CIs are not corrected for testing eight statistics. The $1$-step teacher is also not a drop-in proxy: with classifier-free guidance it needs two forward passes per candidate on SD3.5, and it does not rank the finer statistics. Whether it could replace the proxy on coarse axes when no distilled sibling exists is untested.

\begin{table}[t]
  \centering
  \footnotesize
  \caption{%
    \textbf{Paired null controls} (Fig.~\ref{fig:transfer}~(b)). Mean per-prompt Spearman $\rho$ against the full teacher ($50$ FLUX / $28$ SD3.5 steps), $12$ GenEval prompts $\times$ $32$ shared noises at $1024^{2}$. \emph{Raw}: global mean of $z_T$. \emph{T-1}: undistilled teacher, $1$ step (teacher guidance). \emph{P-1} / \emph{P-4}: distilled proxy, $1$ / $4$ steps. Bold: $95\%$ CI from a bootstrap over noises excludes $0$ ($B\!=\!2000$, not corrected for the eight statistics).%
  }
  \label{tab:appendix:paired-null}
  \begin{tabular}{lcccccccc}
    \toprule
    & \multicolumn{4}{c}{FLUX} & \multicolumn{4}{c}{SD3.5} \\
    \cmidrule(lr){2-5}\cmidrule(lr){6-9}
    Statistic & Raw & T-1 & P-1 & P-4 & Raw & T-1 & P-1 & P-4 \\
    \midrule
    brightness & $0.22$ & $\mathbf{0.68}$ & $\mathbf{0.53}$ & $\mathbf{0.53}$ & $-0.03$ & $\mathbf{0.91}$ & $\mathbf{0.85}$ & $\mathbf{0.86}$ \\
    color temp. & $\mathbf{-0.25}$ & $\mathbf{0.27}$ & $\mathbf{0.50}$ & $\mathbf{0.50}$ & $-0.08$ & $\mathbf{0.50}$ & $\mathbf{0.54}$ & $\mathbf{0.60}$ \\
    compressibility & $-0.21$ & $\mathbf{0.23}$ & $\mathbf{0.46}$ & $\mathbf{0.46}$ & $0.21$ & $\mathbf{0.46}$ & $\mathbf{0.50}$ & $\mathbf{0.51}$ \\
    saturation & $\mathbf{-0.20}$ & $\mathbf{0.39}$ & $\mathbf{0.33}$ & $\mathbf{0.32}$ & $0.04$ & $\mathbf{0.53}$ & $\mathbf{0.47}$ & $\mathbf{0.51}$ \\
    sharpness & $-0.12$ & $\mathbf{0.20}$ & $\mathbf{0.42}$ & $\mathbf{0.38}$ & $0.15$ & $\mathbf{0.36}$ & $\mathbf{0.40}$ & $\mathbf{0.40}$ \\
    contrast & $-0.08$ & $\mathbf{0.35}$ & $\mathbf{0.32}$ & $\mathbf{0.30}$ & $-0.09$ & $\mathbf{0.46}$ & $\mathbf{0.38}$ & $\mathbf{0.40}$ \\
    colorfulness & $-0.08$ & $\mathbf{0.15}$ & $\mathbf{0.20}$ & $\mathbf{0.20}$ & $0.01$ & $\mathbf{0.29}$ & $\mathbf{0.28}$ & $\mathbf{0.33}$ \\
    edge density & $-0.03$ & $0.08$ & $\mathbf{0.23}$ & $\mathbf{0.14}$ & $0.04$ & $0.06$ & $\mathbf{0.24}$ & $\mathbf{0.30}$ \\
    \bottomrule
  \end{tabular}
\end{table}

\subsection{Shared noise across models and learned noise predictors}
\label{sec:appendix:shared-noise}

These controls ask whether the screened statistic is carried by the noise itself. They pass the same pool through both the FLUX and the SD3.5 pairs, with proxies at $k_\mathrm{proxy}\!=\!4$ and the eight statistics only.

\paragraph{Cross-architecture agreement.} Within each pair, the $4$-step proxy ranks the teacher's statistics on these noises ($p\!<\!0.01$ on all eight statistics, column P-4 of Tab.~\ref{tab:appendix:paired-null}). Across architectures, the FLUX and SD3.5 \emph{teachers} rendering the identical $z_T$ agree only weakly: mean $|\rho|\!=\!0.07$ over the eight statistics and $|\rho|\!\le\!0.07$ on seven of them. They share a weak brightness component ($\rho\!=\!0.26$, $95\%$ noise-bootstrap CI $[0.08, 0.40]$, uncorrected for the eight statistics tested). The two proxies share a similar brightness component ($\rho\!=\!0.35$). The statistic a model produces from a given noise is therefore largely, though not entirely, specific to that model.

\paragraph{Learned noise predictors.} To test whether a learned function of the raw noise could replace the proxy, we fit a ridge regressor and a small MLP for each prompt, statistic and model. Their inputs are $38$ hand-crafted features of $z_T$: per-channel means and standard deviations, global moments and spectral summaries, with no prompt conditioning. We use $5$-fold cross-validation over the $32$ noises with teacher statistics as labels. Within a model, the ridge recovers part of the teacher's brightness rank on held-out noises (FLUX $0.39$, SD3.5 $0.55$). Mean $|\rho|$ over the eight statistics is $0.18$ on FLUX and $0.15$ on SD3.5 for the ridge, and $0.20$ and $0.17$ for the MLP. Raw-noise features thus carry some signal once a predictor is trained on a model's own teacher outputs. Across models, a predictor fit on one teacher and evaluated on held-out noises of the other transfers brightness at only $0.15$--$0.18$, not distinguishable from zero at the noise level, and every other statistic stays near zero. Such a predictor needs teacher renders of the target model as labels. The distilled proxy does not.

\paragraph{Scope.} These controls use a $4$-step proxy at $1024^{2}$ on $12$ prompts, unlike the $1$-step, $512^{2}$, $100$-prompt rank-transfer study. The learned-predictor result is exploratory (fixed hyperparameters, no model selection). We do not compare a learned predictor with the proxy as a screening rule inside \textsc{ProxySearch}.

\subsection{High-pass control}
\label{sec:appendix:fourier}

An alternative account of the split is that the statistical verifiers read a fixed low-frequency band that proxy and teacher share. If a verifier depended only on frequencies below a cutoff, high-pass filtering both images above that cutoff would send the verifier to a constant and its proxy--teacher rank correlation to zero. We render a separate FLUX pool ($50$ hand-written prompts $\times$ $32$ noises at $512^{2}$, $1$-step FLUX-schnell proxy and $50$-step FLUX-dev teacher on the same noises). Before scoring, both images pass through an ideal radial high-pass filter computed via FFT in image space. It zeroes the coefficients within $1/8$ of the corner radius ($0.177$ of the per-axis Nyquist frequency, $2.5\%$ of coefficients including DC), and the signed residual is clipped to $[0,255]$ and quantized to $8$ bits. The high-frequency statistics still transfer after filtering (compressibility $0.42$, sharpness $0.29$, edge density $0.20$), as expected for statistics that read high-frequency content by construction. The low-band statistics cannot be read the same way. Removing DC makes the residual zero-mean, and after clipping about $63\%$ of output values are zero. Filtered ``brightness'' then tracks the amplitude of the high-frequency residual rather than image brightness (across $120$ FLUX renders, Spearman $0.97$ with residual RMS and $-0.08$ with the original brightness). The same rectification affects color temperature, colorfulness, saturation and contrast, and the semantic verifiers see out-of-distribution residual images. The control therefore neither supports nor rules out a low-band explanation for the low-band statistics.

\section{Search results: uncertainty, controls and other backbones}
\label{sec:appendix:search-controls}
\label{sec:appendix:robustness}

This section gives the uncertainty on Tab.~\ref{tab:hero}, shows that the statistical gain comes from ranking rather than pool size, and reports the cross-backbone pilot (Sec.~\ref{sec:exp}).

\subsection{Paired-bootstrap CIs for the headline result}
\label{sec:appendix:hero-ci}

Tab.~\ref{tab:appendix:hero-ci} reports the \textsc{ProxySearch}$-$teacher-argmax delta for every metric of Tab.~\ref{tab:hero}. The statistical $\ell_1$ delta excludes zero by a wide margin in all three search objectives (CI upper bounds $\le\!-0.056$). In panel (a), the paired $t$ statistic is $-19.3$ and the Wilcoxon signed-rank $z$ is $-18.5$. The CI on the optimized reward includes zero in every panel, with lower bounds of $-0.5$~pp GE, $-0.03$ PickScore and $-0.001$ HPSv2. In panel (c) the HPSv2 upper bound rounds to $0.000$. The held-out GE CIs of panels (b) and (c) also include zero.

Two held-out deltas have nominal CIs that exclude zero: ImageReward in panel (a) and aesthetic in panel (c). Neither verifier is optimized by any stage. The CIs are not corrected for the metrics and panels tested. The aesthetic delta ($+0.004$ on a mean near $5.35$) is negligible. The ImageReward delta also holds under a prompt-cluster bootstrap, but the median per-cell difference is $0$ (signed-rank $p\!=\!0.088$), so it shows no semantic cost rather than a clear gain.

\begin{table}[ht]
  \centering
  \small
  \caption{%
    \textbf{Paired-bootstrap $95\%$ CIs on \textsc{ProxySearch}$-$teacher-argmax deltas for Tab.~\ref{tab:hero}.} $10{,}000$ resamples, $n\!=\!1106$ paired cells per panel. Bold: the CI excludes zero.%
  }
  \label{tab:appendix:hero-ci}
  \setlength{\tabcolsep}{4pt}
  \begin{tabular}{lcccccc}
    \toprule
    & \multicolumn{2}{c}{(a) GE-driven} & \multicolumn{2}{c}{(b) PickScore-driven} & \multicolumn{2}{c}{(c) HPSv2-driven} \\
    \cmidrule(lr){2-3}\cmidrule(lr){4-5}\cmidrule(lr){6-7}
    Metric & $\Delta$ & $95\%$ CI & $\Delta$ & $95\%$ CI & $\Delta$ & $95\%$ CI \\
    \midrule
    stat.\ $\ell_1$   & $\mathbf{-0.070}$ & $\mathbf{[-0.077,-0.063]}$ & $\mathbf{-0.065}$ & $\mathbf{[-0.072,-0.059]}$ & $\mathbf{-0.062}$ & $\mathbf{[-0.068,-0.056]}$ \\
    GE (pp)           & $+0.9$ & $[-0.5,+2.3]$ & $+0.3$ & $[-1.2,+1.7]$ & $-0.32$ & $[-1.81,+1.18]$ \\
    PickScore         & $+0.02$ & $[-0.03,+0.06]$ & $-0.01$ & $[-0.03,+0.02]$ & $-0.02$ & $[-0.06,+0.01]$ \\
    HPSv2             & $+0.000$ & $[-0.001,+0.001]$ & $+0.000$ & $[-0.001,+0.001]$ & $-0.000$ & $[-0.001,+0.000]$ \\
    CLIP              & $+0.001$ & $[-0.001,+0.003]$ & $-0.001$ & $[-0.002,+0.001]$ & $+0.000$ & $[-0.001,+0.002]$ \\
    ImageReward       & $\mathbf{+0.06}$ & $\mathbf{[+0.02,+0.10]}$ & $-0.01$ & $[-0.03,+0.02]$ & $-0.01$ & $[-0.03,+0.02]$ \\
    aesthetic         & $+0.000$ & $[-0.003,+0.003]$ & $+0.001$ & $[-0.001,+0.004]$ & $\mathbf{+0.004}$ & $\mathbf{[+0.002,+0.006]}$ \\
    \bottomrule
  \end{tabular}
\end{table}

\subsection{More teacher candidates do not tighten the statistic}
\label{sec:appendix:matched-pool}
\label{sec:appendix:teacher-scaling}

\paragraph{Matched candidate pool.} The headline fixes the number of teacher trajectories, so \textsc{ProxySearch} screens $64$ candidates while teacher argmax renders $8$. To separate pool size from ranking on the statistic, we give the teacher-side baselines the same $64$-candidate pool. All rows of Tab.~\ref{tab:appendix:matched-pool} use the same $100$ cells (pilot50 $\times$ $\{\tau_\mathrm{bright}, \tau_\mathrm{dark}\}$) and one $64$-noise pool at FLUX-$1024$. Teacher argmax at $N_t\!=\!8$ uses the first $8$ noises of that pool, and \textsc{ProxySearch} uses $N\!=\!64$, $K\!=\!8$, $k_\mathrm{proxy}\!=\!1$.

Enlarging the teacher pool from $8$ to $64$ does not tighten the statistic under semantic-only selection ($0.304\!\to\!0.306$, paired difference $+0.002$, $95\%$ CI $[-0.006, +0.010]$). The statistical gain comes from ranking on the statistic, not from pool size. At equal teacher trajectories on the same pool, \textsc{ProxySearch} tightens $\ell_1$ by $25\%$ relative to teacher argmax (paired difference $-0.075$, $95\%$ CI $[-0.098, -0.053]$). With the full $64$-candidate teacher pool, the composite reaches a tighter statistic than \textsc{ProxySearch} ($0.143$ vs.\ $0.230$) at $6.9\times$ the total NFE ($8\times$ the teacher NFE). \textsc{ProxySearch} recovers about half of the composite's $\ell_1$ reduction at the teacher cost of $N_t\!=\!8$.

\begin{table}[ht]
  \centering
  \small
  \caption{%
    \textbf{Matched candidate pool} (FLUX-$1024$, $100$ cells, one shared $64$-noise pool). GE differences among the $N_t\!=\!64$ rows and \textsc{ProxySearch} are within noise. On these cells every $\lambda\!\in\!\{0.1, 0.5, 1, 2\}$ selects the same composite image. Selection used the in-process GenEval re-implementation, and GE re-scores the selected images with the official pipeline.%
  }
  \label{tab:appendix:matched-pool}
  \begin{tabular}{lcccc}
    \toprule
    Method & teacher candidates & total NFE (teacher) & stat.\ $\ell_1$ & GE (\%) \\
    \midrule
    Teacher argmax & $8$ & $400$ ($400$) & $0.304$ & $88.0$ \\
    Teacher argmax & $64$ & $3200$ ($3200$) & $0.306$ & $89.0$ \\
    Composite reward ($\lambda\!=\!0.1$) & $64$ & $3200$ ($3200$) & $\mathbf{0.143}$ & $89.0$ \\
    \textbf{\textsc{ProxySearch}} ($N\!=\!64$, $K\!=\!8$) & $8$ & $\mathbf{464}$ ($400$) & $0.230$ & $89.0$ \\
    \bottomrule
  \end{tabular}
\end{table}

\paragraph{Teacher-pool scaling.} We also rerun teacher argmax at $N_t\!=\!16$ on the first $200$ GenEval prompts in file order ($80$ single-object, $99$ two-object and $21$ counting prompts, $n\!=\!400$ cells). The $N_t\!=\!8$ and \textsc{ProxySearch} rows of Tab.~\ref{tab:appendix:nt-scaling} are the headline run restricted to these cells. The $N_t\!=\!16$ arm is a separate run with an independent noise draw, so it is paired with the other rows by prompt and target but not by noise. Doubling the teacher pool raises GE by $1.25$~pp ($93.75\%\!\to\!95.0\%$) but leaves statistical $\ell_1$ unchanged ($0.298\!\to\!0.296$) at twice the teacher NFE ($400\!\to\!800$). \textsc{ProxySearch} at $K\!=\!8$ matches teacher argmax at $N_t\!=\!8$ on GE ($93.75\%$) and reaches $\ell_1$ $0.226$ at $464$ NFE, $42\%$ fewer than the doubled baseline.

\begin{table}[ht]
  \centering
  \small
  \caption{%
    \textbf{Teacher-pool scaling} (FLUX-$1024$, first $200$ GenEval prompts $\times$ $2$ targets, $n\!=\!400$ cells). The $N_t\!=\!16$ row uses an independent noise draw. GE is higher than in Tab.~\ref{tab:hero}~(a) because these prompts contain only the single-object, two-object and counting categories.%
  }
  \label{tab:appendix:nt-scaling}
  \begin{tabular}{lccc}
    \toprule
    Method & GE acc.\ ($\%$) & stat.\ $\ell_1$ & total NFE \\
    \midrule
    Teacher argmax ($N_t{=}8$)                & $93.75$ & $0.298$ & $400$ \\
    Teacher argmax ($N_t{=}16$)               & $\mathbf{95.00}$ & $0.296$ & $800$ \\
    \textbf{\textsc{ProxySearch} ($K{=}8$, $1$-step proxy)}
                                              & $93.75$ & $\mathbf{0.226}$ & $\mathbf{464}$ \\
    \bottomrule
  \end{tabular}
\end{table}

\section{User study}
\label{sec:appendix:human-study}

This section gives the design, results and ethics disclosures of the forced-choice study behind the human judgments of Sec.~\ref{sec:exp:alternatives}. The study has two main arms (A and B) and a separate no-look pilot (C), with no learned reward model in the loop. Annotators were unpaid lab volunteers. Tab.~\ref{tab:appendix:human-study} summarizes the three arms.

\begin{table}[t]
  \centering
  \small
  \caption{%
    \textbf{User-study arms.} Rate: share of judgments choosing the \textsc{ProxySearch} image. Two-way: annotator$\times$item bootstrap, which accounts for repeated judgments. Arm C has one judgment per pair.%
  }
  \label{tab:appendix:human-study}
  \resizebox{\linewidth}{!}{%
  \begin{tabular}{llcccccc}
    \toprule
    Arm & Question & Annotators & Pairs & Judgments & Rate (\%) & Wilson $95\%$ CI & Two-way $95\%$ CI \\
    \midrule
    A (look named) & which image better fulfils the request & $20$ & $144$ & $600$ & $71.7$ & $[67.9, 75.1]$ & $[63.7, 79.1]$ \\
    B (look named) & which image looks like the genuine, unedited photograph & $20$ (same) & $97$ & $400$ & $84.2$ & $[80.4, 87.5]$ & $[76.2, 91.1]$ \\
    C (no look) & which image is preferred overall as a photograph & $5$ & $180$ & $180$ & $48.9$ ($88/180$) & $[41.7, 56.1]$ & n/a \\
    \bottomrule
  \end{tabular}%
  }
\end{table}

\paragraph{Stimuli.} All images are FLUX.1-dev teacher renders at $1024^{2}$, shown side by side after downscaling to at most $512$ px (JPEG). Left/right placement was randomized per pair. The \textsc{ProxySearch} image is the GenEval argmax over teacher renders of the top $8$ of $64$ one-step FLUX-schnell candidates by statistical $\ell_1$ ($464$ NFE). The teacher-argmax image is the GenEval argmax over $8$ teacher candidates ($400$ NFE). A few items come from the $512^{2}$ pilot of Tab.~\ref{tab:method-ablation}, where teacher compute is not matched (Arm A $4$/$144$, Arm B $2$/$97$, Arm C $15$/$180$). Prompts are GenEval prompts under six requested looks with fixed numeric targets: ``bright, high-contrast'' ($\tau_\mathrm{bright}$), ``dark, moody'' ($\tau_\mathrm{dark}$), warm and cool (color temperature $0.5$ and $-0.1$), and high and low sharpness (Laplacian variance $1000$ and $200$). The warm, cool and sharpness items come from separate $50$-prompt runs at the headline operating point that appear in no other table.

\paragraph{Arm A (request fulfillment).} Annotators read the prompt and the named look (e.g., \emph{A user asked for ``\{prompt\}'', with this look: dark and moody, with low-key lighting. Which image better fulfils the request?}) and judged the look and overall image quality together. We kept only pairs in which, on the displayed images, the \textsc{ProxySearch} image is shifted further in the requested direction than the teacher-argmax image by a fixed margin (brightness $\ge\!0.04$, color temperature $\ge\!0.06$, or a sharpness ratio $\ge\!1.25$). All $144$ pairs ($105$ prompts) satisfy this, so results are conditional on \textsc{ProxySearch} having shifted the statistic. Neither image is required to reach $\tau$. Both images pass the in-process GenEval check equally often ($117$ of $144$ each). Twenty annotators gave $600$ judgments ($2$--$5$ per pair). The arm measures joint content-plus-look fulfillment, not semantic superiority. Rates differ strongly by look: dark $88.4\%$, warm $85.7\%$, cool $82.0\%$, high sharpness $70.1\%$, bright/high-contrast $48.1\%$ and low sharpness $48.2\%$. For high sharpness, the two-way bootstrap interval reaches down to $49.1\%$, and a two-way cluster-robust interval is $[54.0, 86.3]$. We have no explanation for the two looks at chance. In both, the two images rarely both reach the target.

\paragraph{Arm B (perceived naturalness).} Annotators were told that both images had the described look, that one was generated that way and the other was an ordinary photo edited afterwards, and were asked which looks like the genuine, unedited photograph. In fact both images are generated. The comparator is the teacher-argmax image post-processed toward the target: the unclamped brightness/contrast shift of Tab.~\ref{tab:method-ablation} for bright and dark looks, a red/blue gain for warm and cool looks, and an unsharp mask or Gaussian blur for the sharpness looks. The last three operations appear nowhere else in the paper. Pairs were only required to be comparable on the targeted statistic, within a fixed tolerance. On tone and color the post-processed image is usually the closer of the two to $\tau$. The same twenty annotators, with Arm A and Arm B items interleaved in individually shuffled orders, gave $400$ judgments on $97$ pairs. Because both images are generated, the arm measures perceived photographic naturalness under a ``one image was edited'' framing, not realism against real photographs. Participants were not debriefed about this framing afterwards. The effect is not significant for low sharpness ($5$ pairs).

\paragraph{Arm C (no-look pilot).} In a separate pilot, five annotators each judged $36$ distinct pairs ($180$ judgments, one per pair). They were asked which image they prefer overall as a photograph. No look was named, no prompt was shown and no manipulation check was applied. Arm C raters saw the same intro page as Arms A and B, which mentions requested looks and edited images, so the pilot is not fully unprimed. The pilot is dominated by bright and dark items ($147$ of $180$ pairs). Rates differed across the five disjoint annotator blocks ($p\!=\!0.007$), and with disjoint item sets this cannot be attributed to annotators rather than items. In this small pilot, neither method is preferred overall.

\paragraph{Procedure and exclusions.} Studies ran on a self-hosted web page. Answers were given by click or arrow key, with no tie option and no time limit. The intro page stated that some questions ask which image better fulfills a requested look, others which image is the genuine, unedited photograph, and that there are no right or wrong answers. All three arms used this intro page. Median response time was about $2$~s per pair. The manipulation check was introduced after an initial batch of $168$ responses had been collected on an earlier stimulus set. That set was regenerated, and those responses are archived and excluded. Excluding the $15$ final judgments on items that the same annotator ID had seen in the earlier batch leaves the Arm A and B rates essentially unchanged ($71.65\%$ and $84.09\%$). No final response was dropped.

\section{Limitations}
\label{sec:appendix:limits}

\paragraph{When \textsc{ProxySearch} does not help.}\label{sec:appendix:when-useful}
\textsc{ProxySearch} needs a statistical target on axes the proxy ranks. We expect little benefit (i) on weakly transferring axes such as edge density, or after teacher reward post-training that attenuates transfer; (ii) without a statistical target, since the proxy is not a usable semantic ranker (Tab.~\ref{tab:appendix:proxy-hf-ceiling}); (iii) for targets that no noise in the pool can reach, since the filter guarantees no tolerance; (iv) when proxy renders collapse across noises~\citep{harrington2026never}; and (v) when $N$ is close to $K$, where the method reduces to teacher argmax.

\paragraph{Scope of the evidence.} ``Statistics'' and ``semantics'' are descriptive labels for the eight statistics and three semantic scorers we measure, not a typology, and a new verifier must be measured before it is assigned to either family. The proxy is a cheap ranker, not a faithful one: its rank agreement peaks at $0.80$ (Tab.~\ref{tab:rho}), which limits how tightly screening can hit the target. Beyond the FLUX headline, our comparisons are single-run pilots, mostly without CIs, and matched teacher trajectories are not matched total compute (App.~\ref{sec:appendix:compute}). The null controls do not pin down a single causal mechanism. Finally, the workflows that motivate the task are motivation only: we have not trained a downstream model on statistic-balanced data.




\end{document}